\documentclass[10pt,twocolumn,letterpaper]{article}

\usepackage{iccv}
\usepackage{times}
\usepackage{epsfig}
\usepackage{graphicx}
\usepackage{amsmath}
\usepackage{amssymb}
\usepackage{listings}
\usepackage[dvipsnames]{xcolor}

\definecolor{codegreen}{rgb}{0,0.6,0}
\definecolor{codegray}{rgb}{0.5,0.5,0.5}
\definecolor{codepurple}{rgb}{0.58,0,0.82}
\definecolor{backcolour}{rgb}{0.95,0.95,0.92}

\lstdefinestyle{mystyle}{
  backgroundcolor=\color{backcolour},
  commentstyle=\color{codegreen},
  keywordstyle=\color{magenta},
  numberstyle=\tiny\color{codegray},
  stringstyle=\color{codepurple},
  basicstyle=\ttfamily\footnotesize,
  breakatwhitespace=false,         
  breaklines=true,                 
  captionpos=b,                    
  keepspaces=true,                 
  numbers=left,                    
  numbersep=5pt,                  
  showspaces=false,                
  showstringspaces=false,
  showtabs=false,                  
  tabsize=2
}

\definecolor{cvprblue}{rgb}{0.21,0.49,0.74}
\usepackage[pagebackref,breaklinks,letterpaper,colorlinks,citecolor=cvprblue,bookmarks=false]{hyperref}

\usepackage[capitalize]{cleveref}
\crefname{section}{Sec.}{Secs.}
\Crefname{section}{Section}{Sections}
\Crefname{table}{Table}{Tables}
\crefname{table}{Tab.}{Tabs.}

\definecolor{ourPurple}{HTML}{9673A6}
\definecolor{ourOrange}{HTML}{D79B00}
\definecolor{ourGreen}{HTML}{82B366}
\definecolor{ourRed}{HTML}{B85450}
\definecolor{personColor}{HTML}{0000FF}
\definecolor{bgColor}{HTML}{bed4f3}

\iccvfinalcopy %

\def\iccvPaperID{4949} %

\newcommand{\qed}{\nobreak \ifvmode \relax \else
      \ifdim\lastskip<1.5em \hskip-\lastskip
      \hskip1.5em plus0em minus0.5em \fi \nobreak
      \vrule height0.75em width0.5em depth0.25em\fi}

\makeatletter
\def\thickhline{%
  \noalign{\ifnum0=`}\fi\hrule \@height \thickarrayrulewidth \futurelet
   \reserved@a\@xthickhline}
\def\@xthickhline{\ifx\reserved@a\thickhline
               \vskip\doublerulesep
               \vskip-\thickarrayrulewidth
             \fi
      \ifnum0=`{\fi}}
\makeatother

\makeatletter
\def\thickhline{%
  \noalign{\ifnum0=`}\fi\hrule \@height \thickarrayrulewidth \futurelet
   \reserved@a\@xthickhline}
\def\@xthickhline{\ifx\reserved@a\thickhline
               \vskip\doublerulesep
               \vskip-\thickarrayrulewidth
             \fi
      \ifnum0=`{\fi}}
\makeatother

\newlength{\thickarrayrulewidth}
\usepackage{booktabs}
\usepackage{balance}
\usepackage{multirow}
\usepackage{overpic}
\usepackage{cite}
\usepackage[font=small,labelfont=bf]{caption}
\usepackage{color}
\usepackage{pifont}
\makeatletter
\@namedef{ver@everyshi.sty}{}
\makeatother
\usepackage{tikz}
\usetikzlibrary{patterns}

\usepackage{colortbl}
\usepackage{pgfplots}
\pgfplotsset{compat=1.17}
\usepackage{tabularx}
\usepackage{arydshln}
\usepackage{amssymb}%
\usepackage{pifont}%

 \usepackage{amsmath}
\usepackage{MnSymbol}%
\usepackage{wasysym}%
\usepackage{enumitem}
\usepackage{wrapfig}
\usepackage{xspace}
\usepackage{tabu}
\usepackage[super]{nth}
\usepackage{subcaption}
\usepackage{scalefnt}
\usepackage{dsfont}
\usepackage{graphicx,calc}

\definecolor{darkgreen}{RGB}{0,153,51}
\definecolor{linkgreen}{RGB}{52,130,48}
\definecolor{LightCyan}{rgb}{0.87,0.92,0.96}
\definecolor{m_green}{RGB}{233, 254, 187}
\definecolor{m_orange}{RGB}{255, 212, 121}
\definecolor{m_red}{RGB}{255, 190, 188}
\definecolor{m_violet}{RGB}{215, 131, 255}
\definecolor{m_blue}{RGB}{186, 234, 255}
\definecolor{m_brown}{RGB}{255,212,120}
\definecolor{m_lightgreen}{RGB}{212,251,122}
\definecolor{notetext}{rgb}{0.7,0,0}

\definecolor{model_pink}{RGB}{235, 106, 164}
\definecolor{model_orange}{RGB}{250, 194, 122}
\definecolor{model_green}{RGB}{164, 210, 162}
\definecolor{model_gray}{RGB}{120, 120, 120}
\definecolor{model_yellow}{RGB}{251, 231, 171}
\definecolor{model_purple}{RGB}{190, 146, 211}

\newcommand{\reffig}[1]{Fig.\,\ref{fig:#1}}
\newcommand{\reftab}[1]{Tab.\,\ref{tab:#1}}

\usepackage{amssymb}%
\usepackage{pifont}%
\newcommand{\cmark}{\ding{51}}%
\newcommand{\xmark}{\ding{55}}%

\newcommand{\ColorMapCircle}[1]{\textcolor{#1}{\ding{108}}}

\newcommand\blfootnote[1]{%
  \begingroup
  \renewcommand\thefootnote{}\footnote{#1}%
  \addtocounter{footnote}{-1}%
  \endgroup
}

\def\eg{\emph{e.g.}\@\xspace} 
\def\ie{\emph{i.e.}\@\xspace} 
 
 \def\vs{\emph{vs.}\@\xspace}
\def\etal{\emph{et al.}\@\xspace}

\newcommand{\circlenum}[1]{{\textcircled{\scriptsize{#1}}}}
\newcommand{\parag}[1]{\vspace{-6px} \vskip8pt \noindent \textbf{#1}}

\newcommand{\name}{Human3D}

\newcolumntype{Y}{>{\centering\arraybackslash}X}
\newcolumntype{Z}{>{\raggedleft\arraybackslash}X}

\newcommand{\colorsquare}[1]{\hspace{-1pt}{\color{#1}$\blacksquare$}\hspace{-6.0pt}$\square$}

\usepackage{array}
\newcolumntype{P}[1]{>{\centering\arraybackslash}p{#1}}
\newcolumntype{M}[1]{>{\centering\arraybackslash}m{#1}}

\definecolor{background}{RGB}{226, 226, 226}
\definecolor{head}{RGB}{210, 78, 142}
\definecolor{rightArm}{RGB}{255, 176, 0}
\definecolor{leftArm}{RGB}{228, 162, 227}
\definecolor{rightForeArm}{RGB}{90, 64, 210}
\definecolor{leftForeArm}{RGB}{243, 232, 88}
\definecolor{rightHand}{RGB}{158, 143, 20}
\definecolor{leftHand}{RGB}{192, 100, 119}
\definecolor{torso}{RGB}{100, 143, 255}
\definecolor{hips}{RGB}{129, 103, 106}
\definecolor{rightUpLeg}{RGB}{243, 115, 68}
\definecolor{leftUpLeg}{RGB}{152, 200, 156}
\definecolor{rightLeg}{RGB}{149, 192, 228}
\definecolor{leftLeg}{RGB}{152, 78, 163}
\definecolor{rightFoot}{RGB}{129, 0, 50}
\definecolor{leftFoot}{RGB}{76, 134, 26}

\newlength\myheight
\newlength\mydepth
\settototalheight\myheight{Xygp}
\newcommand*\restrca[1]{%
  \settototalheight\myheight{Xygp}%
  \settodepth\mydepth{Xygp}%
  \raisebox{-\mydepth+1pt}{\includegraphics[height=\myheight, trim={20pt 5pt 20pt 4pt},clip]{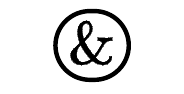}}%
}%

\newcommand*\humanquery[1]{%
  \settototalheight\myheight{Xygp}%
  \settodepth\mydepth{Xygp}%
  \raisebox{-\mydepth+1pt}{\includegraphics[height=\myheight, trim={18pt 0pt 18pt 0pt},clip]{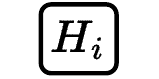}}%
}%

\newcommand*\partquery[1]{%
  \settototalheight\myheight{Xygp}%
  \settodepth\mydepth{Xygp}%
  \raisebox{-\mydepth}{\includegraphics[height=\myheight, trim={18pt 0pt 18pt 0pt},clip]{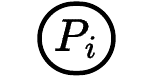}}%
}%

\newcommand*\dotsymbol[1]{%
  \settototalheight\myheight{Xygp}%
  \settodepth\mydepth{Xygp}%
  \raisebox{-\mydepth}{\includegraphics[height=\myheight, trim={50pt 0pt 50pt 0pt},clip]{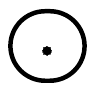}}%
}%

\newcommand*\tausymbol[1]{%
  \settototalheight\myheight{Xygp}%
  \settodepth\mydepth{Xygp}%
  \raisebox{-\mydepth}{\includegraphics[height=\myheight, trim={45pt 0pt 45pt 0pt},clip]{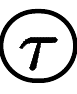}}%
}%

\newcommand*\sigmoidsymbol[1]{%
  \settototalheight\myheight{Xygp}%
  \settodepth\mydepth{Xygp}%
  \raisebox{-\mydepth}{\includegraphics[height=\myheight, trim={60pt 0pt 60pt 0pt},clip]{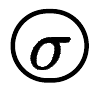}}%
}%

\newcommand{\smallbigcirc}[1]{%
  \vcenter{\hbox{\scalebox{0.77778}{$\m@th#1\bigcirc$}}}%
}

\begin{document}

\title{3D Segmentation of Humans in Point Clouds with Synthetic Data}

\author{Ay\c{c}a Takmaz$^{*1}$ \hspace{5px}
Jonas Schult$^{*2}$ \hspace{5px}
Irem Kaftan$^{\dagger1}$ \hspace{5px}
Mertcan Ak\c{c}ay$^{\dagger1}$ \hspace{5px}
Bastian Leibe$^2$ \hspace{5px}
Robert Sumner$^1$\\
Francis Engelmann$^{1,3}$ \hspace{10px} Siyu Tang$^1$
\\
{\small
$^1$ETH Z\"urich, Switzerland
\hspace{15px}
$^2$RWTH Aachen University, Germany
\hspace{15px}
$^3$ETH AI Center, Switzerland
}
\vspace{1px}\\
\\\\
}

\twocolumn[{%
\renewcommand\twocolumn[1][]{#1}%
\maketitle
\vspace{-12pt}
\vspace{-42pt} 
\begin{center}
    \begin{overpic}[width=1.0\textwidth]{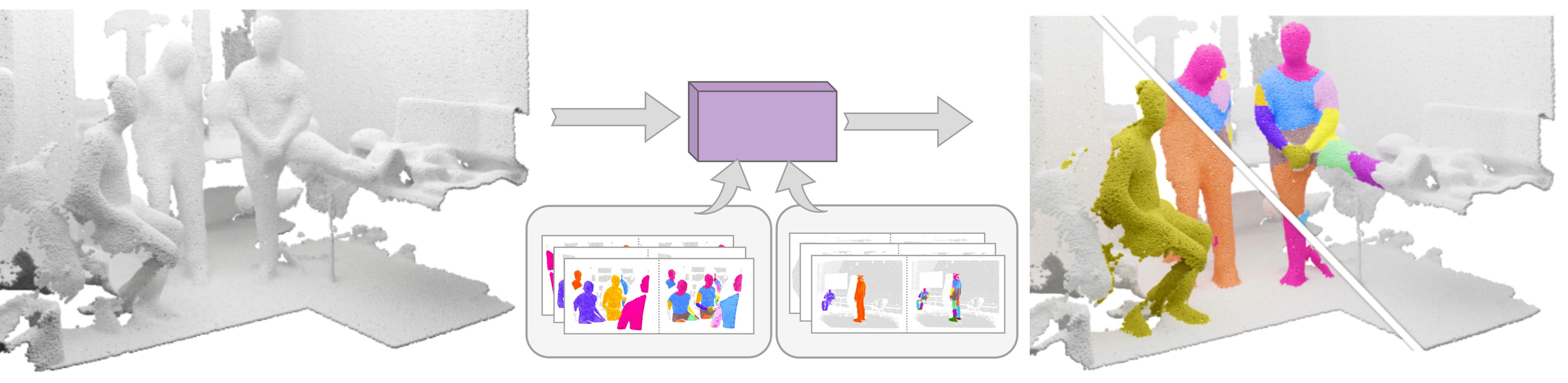}  %
    \put(20,22){\small \emph{Input Point Cloud}}
    \put(45.,16){\small \name{}}
    \put(40.5, 22.5){\small \href{https://human-3d.github.io}{\texttt{human-3d.github.io}}}
    \put(87.25,21.5){\small \emph{Multi-Human}}
    \put(89.5,19.5){\small \emph{Body-Part}}
    \put(87.25,17.5){\small \emph{Segmentation}}
    \put(78.5,4.25){\small \emph{Instance}}
    \put(78.5,2.25){\small \emph{Segmentation}}
    \put(34.5,9.9){\scriptsize \emph{Synthetic Training Data}}
    \put(52,9.9){\scriptsize \emph{Real Training Data}}
    \put(37,2.3){\tiny \color{Plum} \bf{instances}}
    \put(43,2.3){\tiny \color{orange} \bf{body parts}}
    \put(53,2.3){\tiny \color{Plum} \bf{instances}}
    \put(59,2.3){\tiny \color{orange} \bf{body parts}}
    \end{overpic}
\vspace{-18px}
\captionof{figure}{
We propose \name{}, the first end-to-end model for 3D multi-human body-part segmentation in point clouds.
Motivated by the lack of diverse and accurately labeled 3D human datasets,
we generate synthetic training data of virtual humans in realistic 3D indoor scenes and demonstrate its potential in combination with pseudo-labels on real data. 
Above, we show an in-the-wild example of our model that is trained on synthetic data and real Kinect depth data, and tested on a reconstructed point cloud scanned with an iPhone LiDAR sensor.}
\label{fig:teaser}
\vspace{-3px}
\end{center}
}]

\maketitle

\begin{abstract}
\vspace{-7px}
Segmenting humans in 3D indoor scenes has become increasingly important with the rise of human-centered robotics and AR/VR applications.
To this end, we propose the task of joint 3D human semantic segmentation, instance segmentation and multi-human body-part segmentation.
Few works have attempted to directly segment humans in cluttered 3D scenes, which is largely due to the lack of annotated training data of humans interacting with 3D scenes. 
We address this challenge and propose a framework for generating training data of synthetic humans interacting with real 3D scenes.
Furthermore, we propose a novel transformer-based model, \name{}, which is the first end-to-end model for segmenting multiple human instances and their body-parts in a unified manner. 
The key advantage of our synthetic data generation framework is its ability to generate diverse and realistic human-scene interactions, with highly accurate ground truth. 
Our experiments show that pre-training on synthetic data improves performance on a wide variety of 3D human segmentation tasks.
Finally, we demonstrate that Human3D outperforms even {task-specific} state-of-the-art 3D segmentation methods.

\end{abstract}

\vspace{-13px}

\iccvsection{Introduction}
\vspace{-2px}

\label{sec:introduction}

\blfootnote{$^{*,\dagger}$ indicate equal contribution.}
In this work, we address the task of segmenting humans in point clouds. In particular, we focus on 3D semantic segmentation (humans \vs background), 3D instance segmentation (masking multiple humans) and 3D multi-human body-part segmentation (segmenting human instances together with their body parts) as shown in \cref{fig:teaser} \emph{(right)}.

As human-centered robotics and embodied AI are becoming more popular,
there has been a growing interest in the development of methods for 2D human segmentation
\cite{rep-parser,
rprcnn,
parsing-rcnn,
mhpv2,
cihp,
maskrcnn, densepose}
and 3D human detection and segmentation
\cite{humanfast, %
humandetect, %
humanpart, %
depthmap,
shotton2011real}.
While image-based methods have inherent limitations in their ability to reason in 3D, existing 3D methods mainly focus on simplified scenarios in which they only consider individual humans with pre-defined foreground segmentation masks and minimal occlusions. Real-life 3D scenarios, however, are typically cluttered, which can lead to strong occlusions when humans interact closely with each other and their environment.

3D segmentation of humans in point clouds (or depth maps) is a critical aspect of perceiving humans in various applications, such as AR/VR and robotics, in which depth sensors are commonly available and heavily used. For such applications, using point clouds has certain advantages. First, point clouds provide accurate scale and geometry, and are robust against illumination changes. Second, in the realm of human-related computer vision, point clouds are less biased towards visual appearance of humans. This can improve model fairness, and ensures better privacy when collecting data of real humans.

\newpage
Although there have been significant advancements in 3D scene understanding methods that operate directly on point clouds and segment indoor objects \cite{mask3d, mix3d, kpconv, minkowski}, these advancements have not yet translated to the task of 3D \emph{human} segmentation due to a lack of annotated humans in popular 3D indoor training datasets \cite{scannet, s3dis, matterport}.
These indoor datasets usually lack diverse scenarios involving interactions between humans and cluttered real-world indoor environments. %
While outdoor datasets \cite{semantickitti, nuscenes} provide labels for pedestrians, they are limited in terms of human poses, actions, and occlusion patterns, making them less practical for indoor applications where humans closely interact with their surroundings. 
More recently, new datasets (BEHAVE\cite{behave}, RICH\cite{rich}, EgoBody\cite{egobody}) provide depth recordings of humans interacting with their surroundings and other people.
They are labeled with pseudo-ground truth human body meshes \cite{smpl, smplx} via multi-view registration processes relying on image segmentation and manual cleaning.
To facilitate the labeling process, these datasets are often limited in terms of scene complexity, the number of people and poses, as well as occlusion and truncation patterns.
Nevertheless, while tedious to annotate, these datasets can serve as realistic pseudo-labels for training 3D human segmentation tasks.

The key issue of recording and labeling real humans in complex indoor scenes is the time-consuming annotation process and thus its limited scalability. %
A promising alternative is \emph{synthesizing} virtual humans as training data.
Synthetic training data contains perfect labels that are impossible to annotate manually, and the creators have full control over dataset variation and diversity. Compared to generating
color images, where it is challenging to render photo-realistic humans \cite{fakeit}, %
generating depth scans of 3D humans in 3D scenes is significantly easier, as the domain gap between real and synthetic point clouds is much smaller.

In this work, we describe a framework for synthesizing virtual humans in realistic environments, and show that it is possible to create synthetic training data that helps to improve 3D human segmentation in-the-wild.
In addition, we propose a novel transformer-based model, called \name{},
that performs a wide variety of 3D human segmentation tasks in a unified manner.
\name{} is the first model that directly addresses 3D multi-human body-part segmentation in point clouds of realistic environments.
\name{} relies on a novel mechanism using \emph{two-level} queries to jointly segment human instance masks and their associated body parts.
Our experiments consistently demonstrate that pre-training models with synthetic data and fine-tuning with real data yields significant improvements over models trained exclusively on real data.
Furthermore, our \name{} model trained for multi-human body-part segmentation achieves superior performance compared to task-specific state-of-the-art models for both 3D semantic and instance segmentation.

\newpage
In summary, our contributions are as follows:
\begin{itemize}[itemsep=1mm, parsep=1pt]
\item \name{}, the first multi-human body-part segmentation model, that operates directly on real-world cluttered indoor 3D scenes.
\vspace{-3px}
\item An approach for generating synthetic data of humans in 3D scenes and its use for synthesizing training data to improve 3D human segmentation.
\vspace{-3px}
\item Manual annotation of 3D human instances on \mbox{EgoBody} \cite{egobody} to evaluate human segmentation tasks.
\vspace{-3px}
\item Extensive analysis showing the benefits of {pre-training} on synthetic data on multiple baselines and tasks. %
\end{itemize}

\iccvsection{Related work}
\label{sec:relatedwork}

\parag{Multi-human parsing (MHP).}
The goal of MHP is to segment multiple human instances along with their body parts.
While well-explored in images \cite{cihp, mhpv2, parsing-rcnn, rprcnn, rep-parser, maskrcnn},
it received less attention in point clouds. Several approaches \cite{parsing-rcnn, rprcnn} are based on Mask R-CNN \cite{maskrcnn} which is one of the most effective methods for 2D instance segmentation. Yang \etal proposed RP~R-CNN \cite{rprcnn} which combines instance segmentation with semantics using a global semantics-enhanced feature pyramid network. While all of these methods require color images and cannot operate on purely geometric data such as point clouds, MHP and multi-human body-part segmentation in 3D are two very related tasks. As RP~R-CNN \cite{rprcnn} defines the state-of-the-art in MHP and is easily adaptable to our task, we consider RP~R-CNN as a natural choice for a strong baseline.

\parag{Segmenting humans in depth scans.}
Several methods have been proposed for detecting humans \cite{humanfast, humandetect} and segmenting humans or body parts in depth scans \cite{humandetect, depthmap, humanpart, shotton2011real}.
Unlike ours, these methods often assume a given human segmentation mask,
are limited to a single or few humans, and cannot handle strong occlusions.
Instead, we focus on segmenting humans and body parts in real 3D scenes with multiple interacting people under strong occlusions.

\parag{3D semantic and instance segmentation.}
The goal of 3D semantic segmentation is to assign a semantic label to each point in a given 3D scene \cite{markov, semlabel, semindoor, fast, fcn, 3dconv, segcloud, s3dis, scnn, minkowski, vmnet, pointnet, pointnet++, pointcnn, superpoint, spidercnn, kpconv, pointwisecnn, pcnn, spatial, neighbors}. %
Instance segmentation further separates multiple objects within the same semantic class \cite{topdown1, topdown2, bottomup1, bottomup3, bottomup4, voting-based1, voting-based2, softgroup, pointgroup, mask3d, chibane2022box2mask, yue2023agile3d, kreuzberg20224d}. %
The field is largely driven by datasets \cite{scannet, s3dis, matterport} which ignore human labels, so these methods usually cannot segment humans.
In this work, {we train state-of-the-art} methods KPConv\cite{kpconv}, MinkowskiUNet \cite{minkowski}, and Mask3D\cite{mask3d} on our proposed data, and compare them on different human segmentation tasks.
Building on \cite{minkowski, mask3d}, we propose the first end-to-end model for 3D multi-human body-part segmentation. In particular, the key idea of \name{} is to use \emph{two-level} queries where the first level represents human masks and the second level represents their associated body parts.

\begin{figure*}[!t]
\centering
\includegraphics[width=1.0\linewidth]{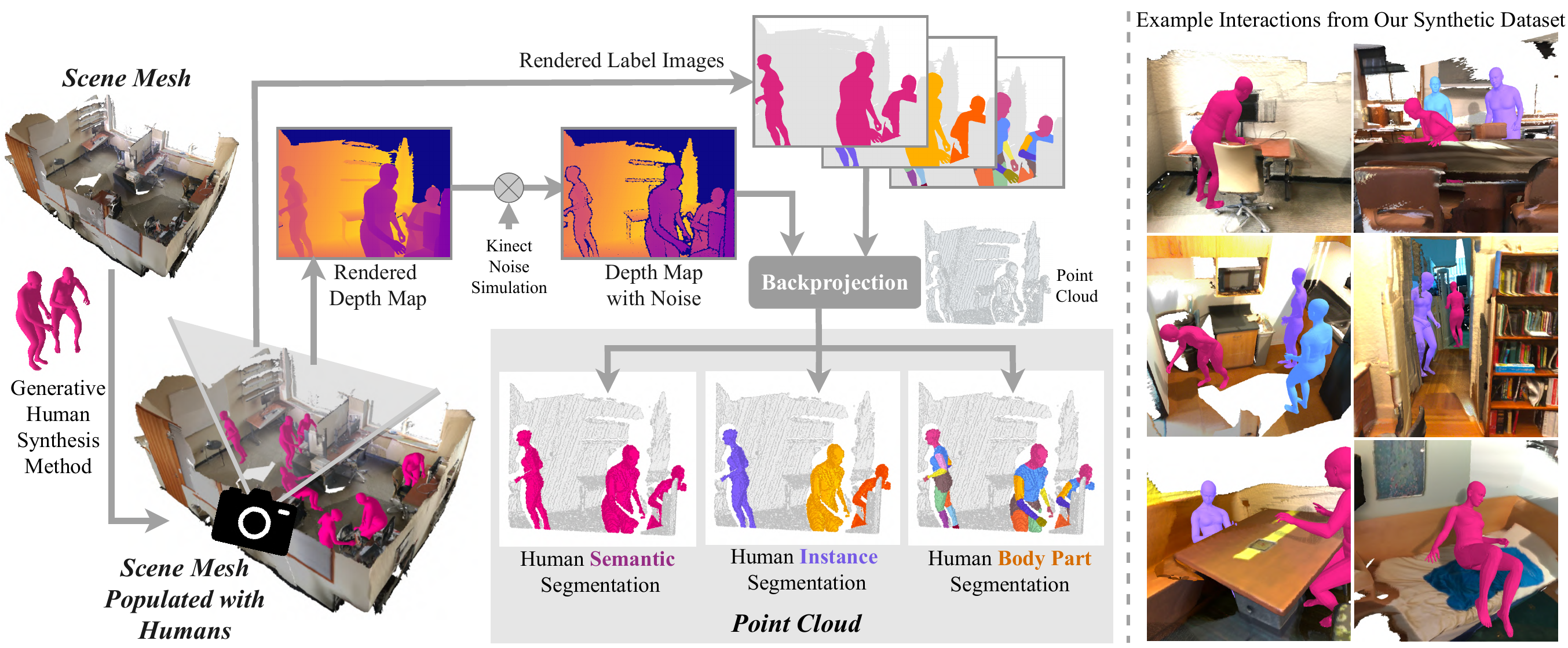}
\caption{\textbf{Synthesizing training scenes.} \textit{Left:} Given a scene mesh from ScanNet\cite{scannet},
we populate it with \emph{synthetic humans} based on PLACE\cite{place}. We then render label maps and depth maps augmented with simulated Kinect noise \cite{simkinect}. Finally, the labels are backprojected to 3D using the synthesized depth maps to obtain highly accurate labels for human semantic, instance, and body-part segmentation. \textit{Right:} Example interactions from our synthetic dataset featuring multiple humans, various occlusion levels and close contact with scene objects.
}
\label{fig:synthetic_data_generation}
\vspace{-10px}
\end{figure*}

\parag{Synthetic data generation.}
Accurately annotating large amounts of data is tedious and occasionally not feasible, \eg human body-part segmentation.
This motivates an emerging trend towards synthesizing training data for various computer vision tasks \cite{agora, tome2019xr, urbanscene, fakeit, surreal, train_synthetic, multihuman_flow, hspace, carla, wang2022humanise}.
SURREAL \cite{surreal} synthesizes 2D humans on top of real color images. However, the synthesized humans are not conditioned on the images, which results in unrealistic renderings.
HSPACE \cite{hspace} is a large-scale dataset of synthetic humans in synthetic indoor and outdoor environments, focusing on generating realistic color images.
HUMANISE \cite{wang2022humanise} is a language-conditioned human motion generator in 3D scenes and provides a dataset of synthetic, moving humans.
Alternative methods \cite{place, posa, psi, tendulkar2022flex} populate 3D scenes with synthetic humans.
PLACE~\cite{place} synthesizes realistic 3D humans with natural poses conditioned on a given 3D scene.
We extend PLACE to generate multiple 3D humans in ScanNet\cite{scannet} scenes and condition the human generation to interactions with specific scene objects (\eg{}, sofa, bed, chair).

\iccvsection{Data Generation}
In Sec.~\ref{sec:synthetic-data-generation}, we describe our framework for generating synthetic training data for human instance and body-part segmentation tasks. Then in Sec.~\ref{sec:real-data-collection}, we describe our real data collection, processing and annotation pipelines.

\iccvsubsection{Synthetic Training Data Generation}
\cref{fig:synthetic_data_generation} illustrates our framework for generating synthetic training data.
It populates real indoor scenes with synthetic humans and automatically generates labeled point clouds with perfect human and body-part labels that are otherwise difficult to obtain by manual labeling.

\label{sec:synthetic-data-generation}
\parag{Populating 3D indoor scenes.}
We populate indoor 3D scenes from ScanNet\cite{scannet}, although our pipeline is suitable for other 3D indoor datasets as well \cite{s3dis, matterport, rio}. To place synthetic humans in a given scene, we base our pipeline on PLACE  \cite{place}, which is a generative human-scene interaction synthesis method.
In order to obtain a large variety of human poses and close human-scene interactions,
we modify \cite{place} to perform instance segmentation-guided human placement.
In our approach, we first identify object categories with which humans can naturally have close contact (\eg{} chairs, tables),
and use 3D object instance labels from ScanNet \cite{scannet} to  select  these objects in the human-scene interaction synthesis process.
We then sample potential interaction objects to generate up to 10 synthetic humans per scene, along with their SMPL-X \cite{smplx} body parameters. The human synthesis approach is scene-aware as it encodes the nearby scene features. Our pipeline enables us to generate humans in various poses while taking human-scene proximity into account for close interactions. Further details about the human synthesis pipeline are in the sup. mat. Sec.~\textcolor{red}{1}. %

\parag{Rendering.} We render depth maps and label images from scene meshes we populate with humans. A virtual camera is placed at the scene center (arithmetic mean of the scene vertex coordinates), and its height is uniformly sampled from $[1.4, 1.6]\mathrm{m}$ to reflect the height of a potential handheld capture device (e.g. mobile phone, tablet). Camera viewing direction is always in parallel to the ground plane (xy-plane) and is rotated around the vertical axis by an amount uniformly sampled within $[0^{\circ}, 360^{\circ})$. Rendered label images include annotations for semantics, instances, and multi-human body-parts (\cref{fig:synthetic_data_generation}, \textit{top}). We capture 40 frames per ScanNet scene, and re-sample the camera pose at each iteration. Further details about camera placement and sampling parameters are provided in the sup. mat. Sec.~\textcolor{red}{1.2}.

\parag{Simulating Kinect noise.}
We further refine the rendered depth maps by simulating Kinect noise using \cite{simkinect} to more closely mimic the depth data from a real Kinect sensor, as we use real Kinect data from EgoBody\cite{egobody} for evaluation (Sec.\,\ref{sec:eval-data-annotation-exp}). This allows us to combine real Kinect data (Sec.\,\ref{sec:real-data-for-training}) and synthetic data for training. In preliminary studies, we found that simulating Kinect noise positively influences the segmentation quality. Please see sup. mat. Sec.~\textcolor{red}{1.3} and Fig.~\textcolor{red}{2} for further details and illustrations. %
\begin{figure}[!t]
\begin{minipage}{0.3\textwidth}
    \includegraphics[height=2.1in]{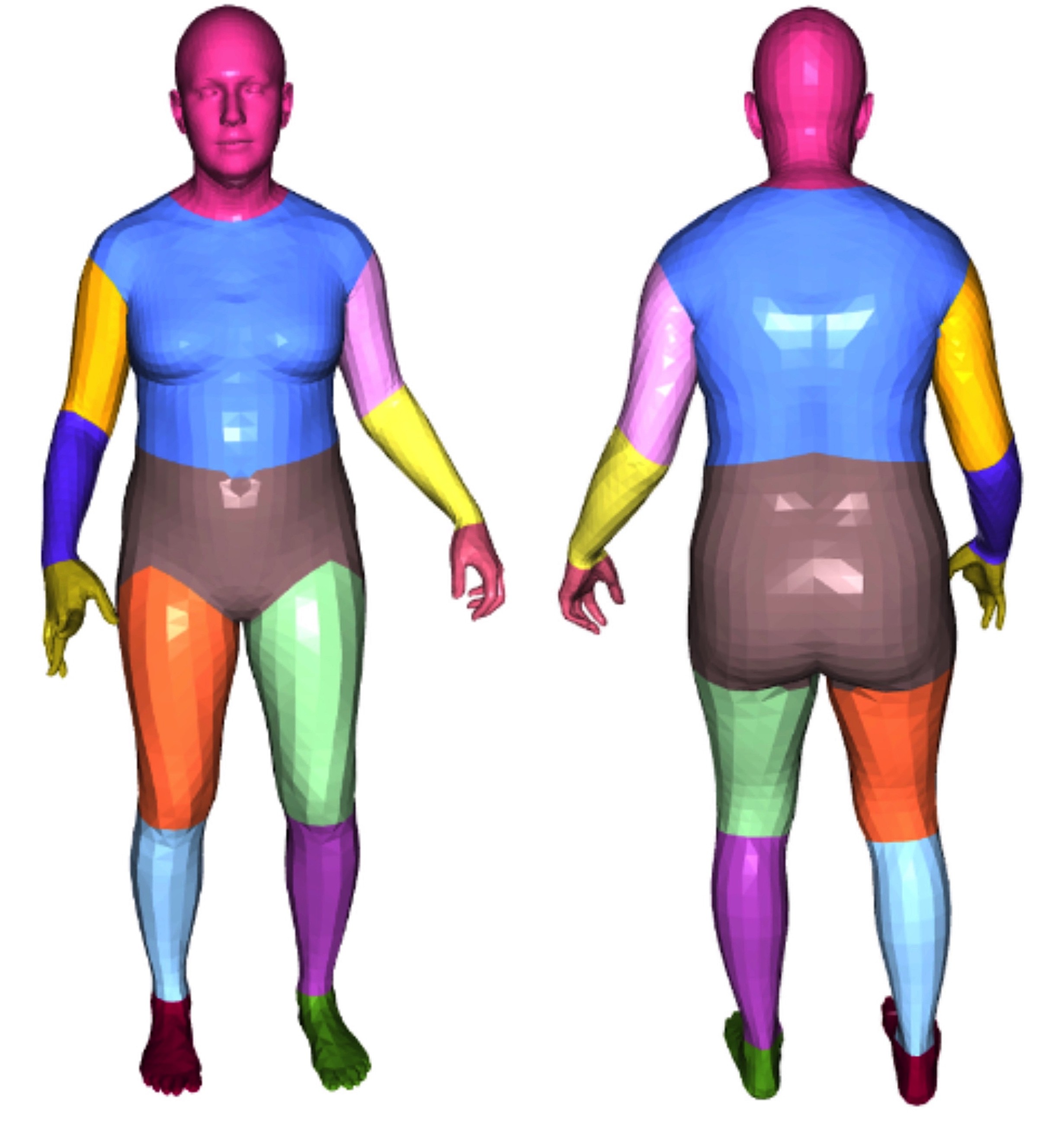}
\end{minipage}
\hfill    
\begin{minipage}{.1\textwidth}
 \footnotesize
    {
        \ColorMapCircle{head}\,Head
        \ColorMapCircle{rightArm}\,RightArm
        \ColorMapCircle{leftArm}\,LeftArm
        \ColorMapCircle{rightForeArm}\,RightForeArm
        \ColorMapCircle{leftForeArm}\,LeftForeArm
        \ColorMapCircle{rightHand}\,RightHand
        \ColorMapCircle{leftHand}\,LeftHand
        \ColorMapCircle{torso}\,Torso
        \ColorMapCircle{hips}\,Hips
        \ColorMapCircle{rightUpLeg}\,RightUpLeg
        \ColorMapCircle{leftUpLeg}\,LeftUpLeg
        \ColorMapCircle{rightLeg}\,RightLeg
        \ColorMapCircle{leftLeg}\,LeftLeg
        \ColorMapCircle{rightFoot}\,RightFoot
        \ColorMapCircle{leftFoot}\,LeftFoot
        }
\end{minipage}
\vspace{-10px}
\caption{
\textbf{Body-parts.} After merging smaller parts into larger ones (e.g. eyes into head), we obtain 15 body-part labels.
}

\label{fig:body_part_schema}
\end{figure}

\begin{figure*}[!t]
  \centering
   \includegraphics[draft=False, width=0.99\linewidth]{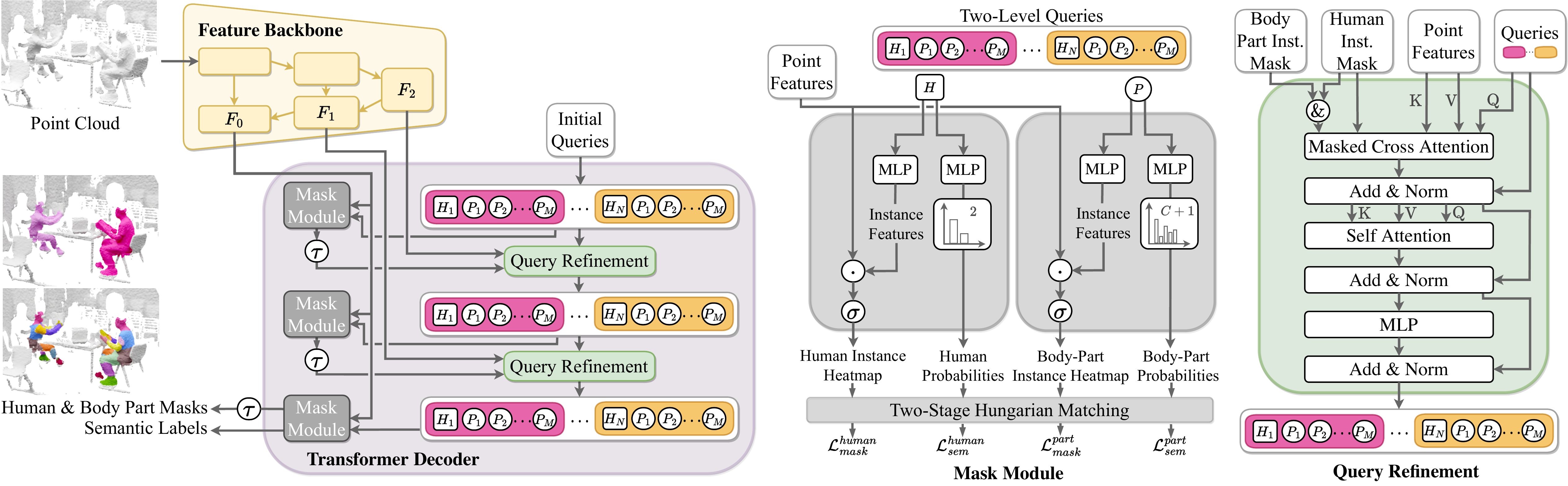}%
\vspace{-6px}
\caption{\textbf{Illustration of the \name{} model.}
Our model consists of a sparse convolutional feature backbone and a transformer decoder \emph{(left)}.
The mask module jointly predicts human instance masks and body-part masks based on two-level queries \emph{(middle)}, which are iteratively refined based on multi-scale point features within the predicted human instance mask \emph{(right)}.
\protect\humanquery{} represents human queries and \protect\partquery{} represents body-part queries. %
\protect\tausymbol{} applies a threshold of 0.5, \protect\sigmoidsymbol{} is the sigmoid function and \protect\dotsymbol{} is the dot product operation.}   %
\label{fig:model}
\vspace{-8px}
\end{figure*}

\parag{Labeled point clouds.} The resulting depth maps and label images are back-projected into 3D space to obtain perfectly labeled partial point clouds. %
We use this pipeline to create a synthetic dataset for human semantic, instance, and multi-human body-part segmentation (MHBPS). For MHBPS, we map the faces of each SMPL-X \cite{smplx} mesh to body-parts according to \cite{meshcapade}, then merge smaller parts into larger ones (\eg{} eyes into head) and obtain 15 body-part classes. Resulting list of body parts is illustrated in Fig.~\ref{fig:body_part_schema}. Please see the sup. mat. Sec.~\textcolor{red}{1.5-1.6} and Tab.~\textcolor{red}{1} for additional details.

\iccvsubsection{Real Data Collection}
\label{sec:real-data-collection}
\subsubsection{Pseudo Training Labels on Real Data}
\label{sec:real-data-for-training}
Besides the synthetic data with perfect labels, we can also use real training data even though it requires expensive and time-consuming capturing processes and it produces less accurate, \ie \textit{pseudo,} labels.
We use the recently released 3D human-scene interaction datasets EgoBody\cite{egobody} and BEHAVE\cite{behave}.
BEHAVE includes sequences of individual humans interacting with a single object in a mostly empty scene.
EgoBody features social interactions between two humans captured in more cluttered static scenes.
Both datasets provide multi-view depth recordings from several Kinect sensors, and carefully fitted SMPL\cite{smpl} or SMPL-X\cite{smplx} human body models. 
We obtain point clouds by back-projecting the Kinect depth to 3D and utilize the fitted body model parameters to obtain 3D human segmentation masks. We obtain body-part labels by selecting scene points within a fixed distance (5 cm) from the fitted body mesh, and assign each point to the closest body-part in the fitted body. Please refer to sup. mat. Sec.~\textcolor{red}{2} for more details. 

\vspace{-6px}
\subsubsection{Manually Refined Evaluation Dataset}
\label{eval-data-annotation}
\vspace{-3px}

Pseudo-ground truth labels for human masks and body parts that were extracted using multi-view fitted body models from EgoBody (as described in Sec.~\ref{sec:real-data-for-training}) can be noisy in certain scenarios such as close-contact interactions with scene objects (\eg sitting on a sofa), loose clothing (\eg wide-legged jeans) or unusual poses (causing a mismatch between the fitted body mesh and real human point cloud). As we cannot rely on noisy pseudo-labels for the evaluation of our model, we created a manually refined evaluation set based on the EgoBody dataset for a rigorous evaluation.

\parag{Splits.} The EgoBody\,\cite{egobody} dataset contains $125$ interaction sequences captured by multiple Kinect cameras. As the original train/validation/test split was created with an aim to separate first-person view subjects (the subject observed by the other subject wearing a head-mounted device) in each sequence, we created a new split such that none of the subjects overlap across splits. Our split consists of 73 training sequences, 11 validation sequences, as well as 38 test sequences, while 3 sequences were removed to ensure a non-overlapping distribution of subjects across splits. 

\parag{Manual refinement.}
For each of the selected 38 test sequences, expert annotators have annotated 8 scenes (point clouds), resulting in a test set consisting of 304 point clouds featuring a large variety of human poses, action types and occlusion levels. The annotation process is performed using a 3D annotation tool \cite{annotation-tool}.  The labeling process is initialized with the noisy pseudo-labels for human instances based on the existing multi-view fitted human meshes (Sec.~\ref{sec:real-data-for-training}). Then, the human instance masks are manually refined by the annotators. Body part label refinement is then guided by the resulting ground-truth human instance masks such that each point in the human mask is assigned to the closest body part in the original fitted body, and each point outside of the refined human mask is removed from the body part mask. Further details are in the sup. mat. Sec.~\textcolor{red}{2}.

\iccvsection{Multi-Human Body-Part Segmentation}

Our approach, \name{}, addresses the task of multi-human body-part segmentation (MHBPS) on 3D point clouds, \ie it detects individual human instances and semantically partitions them into body-parts.
Complex 3D indoor environments, diverse human-object interactions, and close distances between humans make this task challenging.
Not only is it required to correctly segment the body-parts, but it is also needed to correctly associate the body-parts with human instances. This needs capturing well-localized geometric details and high-level semantic context.

Inspired by the success of Mask3D\,\cite{mask3d} for 3D instance segmentation,
we propose a transformer-based model with two dedicated query types: one for humans and one for body-part instances.
We call these \emph{two-level} queries.
This key technical contribution enables the structured differentiation between human-level queries \protect\humanquery{} and body-part-level queries \protect\partquery{} (See Fig.\,\ref{fig:model}).
It is also essential to explicitly tie human masks together with their corresponding body-part masks during training such that body-part queries of one person are not supervised with ground truth masks of another person.
Furthermore, we introduce a two-stage Hungarian matching mechanism, which 
guarantees that each ground truth human and body-part instance has a unique match with a predicted human instance and its associated body-parts.
This matching explicitly enforces that human queries are tied to their respective body-part queries.

\parag{Overview.}
Our \name{} model is illustrated in \cref{fig:model}. 
Our architecture consists of (1) a sparse convolutional feature backbone~(\colorsquare{model_yellow}) implemented as a MinkowskiUNet\,\cite{minkowski},
(2) a query refinement step~(\colorsquare{model_green}) implemented as a masked transformer decoder~(\colorsquare{model_purple})\,\cite{Cheng22CVPR} which iteratively refines human and body-part queries by cross-attending to the multi-resolution hierarchy of the backbone decoder's point features $\{F_i\}_{i=0}^2$, and (3) a mask module~(\colorsquare{model_gray}) which predicts heatmaps for human and body-part instances together with their associated semantic class label. %

\parag{Human and Part Query Types.}
The key technical contribution of this model, compared to prior work \cite{mask3d}, is the two-level query types where each level specializes on one downstream task:
The first level represents the human queries $H_1, ..., H_N$ (shown as \protect\humanquery{} in \cref{fig:model}) which are trained to segment up to $N$ human instances in a scene.
The second level represents the body-part queries $\{ P^{i}_1, ..., P^{i}_M \}_{i=1}^N$ (shown as \protect\partquery{} in \cref{fig:model}).
To each one of the $N$ human queries, we associate $M$ body-part queries.
This explicit modelling of correspondences between $M$ body-part queries and a single corresponding human query, results in two important properties:
(1) We can directly extract the body-part segmentation for each human instance and
(2) during query refinement (\cref{fig:model}, \colorsquare{model_green}),
we enable information flow between human instances and body-parts via self-attention among human and body-part queries.
We therefore update human instance masks based on their predicted body-part masks, and vice versa.
Further, we tie body-parts to their associated human instance, by restricting body-parts to only cross-attend to backbone point features which lie within the corresponding human  mask (\cref{fig:model} \protect\restrca{}, \emph{right}).

\parag{Two-Stage Hungarian Matching.}
\name{} infers $N$ human instances and $N \cdot M$ body-parts during a single feed forward pass of the model.
As these predictions as well as the ground truth targets are unordered, we need to find optimal correspondences between these two sets in order to optimize the model.
Typically, the Hungarian Algorithm\,\cite{Kuhn55hungarian} is deployed to find such optimal correspondences\,\cite{mask3d, Cheng22CVPR, Carion20ECCV}.
However, for MHBPS we cannot simply match human and body-parts independently.
We additionally have to guarantee that both the predicted body-part masks and the human mask are mapped to target body-part masks and target human mask of the same human.
We therefore introduce a two-stage Hungarian matching approach:

In the first stage, we define the assignment cost for a predicted \emph{human} instance $h$ and a target instance $\hat{h}$ as follows:
\begin{equation}
\small
\mathcal{C}_1(h,\hat{h}) = \mathcal{L}^\textrm{human}_\textrm{mask}(h,\hat{h}) + \mathcal{L}^\textrm{human}_{\textrm{sem}}(h,\hat{h})
\end{equation}
The cost for matching human \emph{masks} is a weighted combination of the Dice loss\,\cite{Deng18ECCV} and binary cross-entropy
$\mathcal{L}^\textrm{human}_\textrm{mask} = \lambda_{\text{BCE}}\mathcal{L}_{\textrm{BCE}} + \lambda_{\text{dice}}\mathcal{L}_{\textrm{dice}}$
while the semantic classificaton loss is defined as
$\mathcal{L}^\textrm{human}_\textrm{sem}$=$\lambda_{\text{cl}}\mathcal{L}_{\textrm{CE}_\textrm{cl}}$.
Using the Hungarian Algorithm\,\cite{Kuhn55hungarian}, we find a globally optimal assignment between predicted and ground-truth human instances.
Following\,\cite{Carion20ECCV}, we represent this assignment by a permutation $\sigma \in \mathfrak{S}_N$ which maps the target human instance $\hat{h}^j$ to the predicted human instance $h^{\sigma(j)}$.
We then use this optimal assignment between human masks to match their corresponding \emph{body-parts} $p$ using the following cost matrix:
\begin{equation}
\small
\mathcal{C}_2(p^{\sigma(j)},\hat{p}^j) = \mathcal{L}^\textrm{part}_\textrm{mask}(p^{\sigma(j)},\hat{p}^j) + \mathcal{L}^\textrm{part}_{\textrm{sem}}(p^{\sigma(j)},\hat{p}^j)
\end{equation}
$\mathcal{L}^\textrm{part}_\textrm{mask}$ and $\mathcal{L}^\textrm{part}_\textrm{sem}$ are analogously defined to their human instance counterparts $\mathcal{L}^\textrm{human}_\textrm{mask}$ and $\mathcal{L}^\textrm{human}_\textrm{sem}$.
After establishing correspondences between human masks and their corresponding body-parts, we optimize all auxiliary predictions after each of the $L$ query refinement steps:
\begin{equation}
\mathcal{L} = \Sigma_l^L ~ \mathcal{L}^{\textrm{human},l}_{\textrm{mask}}  + \mathcal{L}^{\textrm{human}, l}_{\textrm{sem}} + \mathcal{L}^{\textrm{part},l}_{\textrm{mask}}  + \mathcal{L}^{\textrm{part}, l}_{\textrm{sem}}
\end{equation}
This loss enforces that human masks as well as their body-part masks are matched to the same ground truth human.

We provide an outline of the Two-Stage Hungarian Matching algorithm in Listing\,\ref{lst:two_stage}.

\begin{lstlisting}[language=Python, float, caption=Two-Stage Hungarian Matching Algorithm., label={lst:two_stage}, belowskip=-2pt]
def two_stage_matching(h_mask, h_prob, p_mask, 
                       p_prob, h_gt, p_gt):
  # human-level: h_mask, h_prob and GT h_gt
  # part-level: p_mask, p_prob and GT p_gt

  # 1-stage: human-level predictions <-> GT
  h_indx, loss = Hungarian(h_mask, h_prob, h_gt)
  L_total = loss
  
  # for each (pred, gt) matched human instance
  for (pred_i, gt_j) in h_indx:
    mask = p_mask[pred_i]
    prob = p_prob[pred_i]
    gt = p_gt[gt_j]
    
    # 2-stage: part-level predictions <-> GT
    _, p_loss = Hungarian(mask, prob, gt)
    
    L_total += p_loss 
  return L_total
\end{lstlisting}

\parag{Extracting body-part segmentations.}
\name{} represents body-parts as \emph{instances}.
We therefore now describe how we merge these body-part instances to obtain a semantic body-part segmentation for each human instance.
First, we restrict body-parts to lie within their corresponding human instance masks, \ie points of body-parts outside the human mask are set to background.
Second, for each point in the human mask, we obtain the semantic body-part label of the body-part instance mask with the highest confidence. If the highest confidence is below $10\%$, we ignore the prediction and assign the point to background.

\iccvsection{Experiments}

In this section, we first compare our \name{} model with state-of-the-art segmentation methods for 3D point clouds and 2D images (Sec.\,\ref{sec:sota}).
We then provide analysis experiments on occlusions, an ablation study of \name{} and demonstrate the benefits of pre-training with synthetic data (Sec.\,\ref{sec:analysis}).
Finally, we show qualitative results of our approach (Sec.\,\ref{sec:qualitative_results}).
Additional analysis is provided in the supplementary material Sec.~\textcolor{red}{4} and Sec.~\textcolor{red}{5}.

\begin{table*}[!t]
\centering
\setlength{\tabcolsep}{5pt}
\resizebox{\textwidth}{!}{
\begin{tabular}{ll cccccc ccc cc}
\toprule
& & \multicolumn{6}{c}{\em 3D Multi-Human Body-Part Segmentation} & 
\multicolumn{3}{c}{\em 3D Instance Seg.} &
\multicolumn{2}{c}{\em 3D Semantic Seg.}\\
\cmidrule(r){3-8} \cmidrule(r){9-11} \cmidrule(r){12-13}  
Instance segmentation model & Body-part segmentation model & AP$^P_\text{}$ & AP$^P_{50}$  & AP$^P_{25}$ & PCP & PCP$_{50}$ & PCP$_{25}$ &AP$^H_\text{}$ & AP$^H_{50}$ & AP$^H_{25}$ & mIoU$^H$& mIoU$^P$\\
\midrule
MinkUNet \cite{minkowski} (Human) + Cluster & MinkUNet\cite{minkowski}  (Body Part)  & $8.9$ & $34.8$ & $82.5$ & $8.1$ & $30.6$ & $58.5$ & $68.2$ & $83.6$ & $89.4$ & $92.2$ & $50.8$ \\ %
MinkUNet \cite{minkowski} (Body Part) + Cluster &  MinkUNet \cite{minkowski}  (Body Part) & $9.1$ & $36.0$ & $84.7$ & $8.6$ & $32.6$ & $63.7$ & $76.5$ & $87.2$ & $91.1$ & $92.5$ & $51.3$ \\ 
Mask3D \cite{mask3d}  (Human) & MinkUNet \cite{minkowski} (Body Part)   & $5.9$ & $29.9$ & $90.9$ & $9.2$ & $33.9$ & $65.4$ & $95.6$ & $98.7$ & $99.7$ & $97.6$ & $53.3$ \\
Mask3D \cite{mask3d}  (Human) & KPConv \cite{kpconv} (Body Part)    & $25.5$ & $75.8$ & $98.7$ & $24.4$ & $60.3$ & $74.8$ & $95.6$ & $98.7$ & $99.7$ & $97.6$ & $64.5$ \\ 
KPConv \cite{kpconv} (Human) + Cluster & KPConv \cite{kpconv} (Body Part)  & $28.2$ & $74.7$ & $96.3$ & $22.9$ & $58.4$ & $73.1$ & $89.7$ & $95.3$ & $97.0$ & $96.7$ & $63.6$ \\ %
KPConv\cite{kpconv} (Body Part) + Cluster & KPConv \cite{kpconv} (Body Part)  & $28.8$ & $76.2$ & $97.8$ & $23.4$ & $59.4$ & $74.3$ & $89.3$ & $97.6$ & $98.6$ & $96.8$ & $64.3$ \\ %
\arrayrulecolor{black!10}\midrule\arrayrulecolor{black}
Mask-RCNN+DeepLabv3 2D-3D (as in \cite{egobody}) && -- & -- & -- & -- & -- & -- & 61.3 & 97.3 & 99.8 & 87.7 & -- \\
RP R-CNN 2D-3D\,\cite{rprcnn} && $26.8$ & $80.5$ & $97.3$ & $21.8$ & $61.5$ & $77.6$ & $74.6$ & $97.2$ & $97.9$ & $92.1$ & $58.9$ \\
\arrayrulecolor{black!10}\midrule\arrayrulecolor{black}
\name{} (Ours)  &  & $\mathbf{35.8}$ & $\mathbf{93.2}$ & $\mathbf{99.1}$ & $\mathbf{32.6}$ & $\mathbf{73.5}$ & $\mathbf{84.0}$ & $\mathbf{99.1}$ & $\mathbf{100}$ & $\mathbf{100}$ & $\mathbf{98.3}$ & $\mathbf{69.9}$  \\ %
\bottomrule
\end{tabular}
}
\vspace{-8px}
\caption{\textbf{3D Multi-Human Body-Part Segmentation on EgoBody test set.}
Metrics are average precision for body-parts (AP$^P$) and humans (AP$^H$), correctly parsed semantic parts (PCP) and intersection-over-union on humans (IoU$^H$) and parts (IoU$^P$).
Brackets indicate on which segmentation task the baselines are trained.
3D models are pre-trained on synthetic and fine-tuned on real EgoBody data.
}
\vspace{-15px}%
\label{tab:mhp_benchmark}
\end{table*}

\begin{table}[t]
\centering
\setlength{\tabcolsep}{3pt}
\resizebox{\columnwidth}{!}{
\begin{tabular}{lccccc}
\toprule
       & \multicolumn{2}{c}{\multirow{2}{*}{\em Trained on EgoBody}} && \multicolumn{2}{c}{\em Pre-trained on Synthetic} \\
      & \multicolumn{2}{c}{} && \multicolumn{2}{c}{\em Fine-tuned on EgoBody} \\
\cmidrule(r){2-3} \cmidrule(r){5-6}
Model & \cellcolor{gray!10}AP$^H_\text{}$ & AP$^H_{50}$ && \cellcolor{gray!10}AP$^H_\text{}$ & AP$^H_{50}$ \\
\midrule
MinkUNet \cite{minkowski} {\small(Human) + Cluster} & \cellcolor{gray!10}$64.9$  & $79.6$  && \cellcolor{gray!10}$68.2$ \textcolor{darkgreen}{\small{(+$3.3$)}} & $83.6$ \textcolor{darkgreen}{\small{(+$4.0$)}}\\
MinkUNet \cite{minkowski} {\small(Body Part) + Cluster}& \cellcolor{gray!10}$69.1$ & $81.7$ && \cellcolor{gray!10}$76.5$ \textcolor{darkgreen}{\small{(+$7.4$)}} & $87.2$ \textcolor{darkgreen}{\small{(+$5.5$)}}\\ %
KPConv \cite{kpconv} {\small(Human) + Cluster}
 & \cellcolor{gray!10}$85.4$ & $92.2$ & & \cellcolor{gray!10}$89.7$ \textcolor{darkgreen}{\small{(+$4.3$)}}& $95.3$ \textcolor{darkgreen}{\small{(+$3.1$)}}\\ 
KPConv \cite{kpconv} {\small(Body Part) + Cluster}& \cellcolor{gray!10}$86.9$ & $94.4$ && \cellcolor{gray!10}$89.3$ \textcolor{darkgreen}{\small{(+$2.4$)}} & $97.6$ \textcolor{darkgreen}{\small{(+$3.2$)}}\\
\arrayrulecolor{black!10}\midrule\arrayrulecolor{black}
Mask3D \cite{mask3d} & \cellcolor{gray!10}$89.4$ &	$\mathbf{95.4}$ && \cellcolor{gray!10}$95.6$ \textcolor{darkgreen}{\small{(+$6.2$)}} & $98.7$ \textcolor{darkgreen}{\small{(+$3.3$)}}\\
\name{} (Ours) & \cellcolor{gray!10}$\mathbf{90.5}$ &	$95.2$ && \cellcolor{gray!10}$\mathbf{99.1}$ \textcolor{darkgreen}{\small{(+$8.6$)}} & $\mathbf{100}$ \textcolor{darkgreen}{\small{(+$4.8$)}}  \\ 
\bottomrule
\end{tabular}
}  
\vspace{-6px}
\caption{
\textbf{3D Instance Segmentation Scores on EgoBody test.}
We observe that pre-training with synthetic data results in improvements %
by up to $+8.6$\,AP$^H$.
Further, \name{} outperforms task-specialized models (e.g. Mask3D) by at least $+3.5$\,AP$^H$.
}
\label{tab:inst_seg_benchmark}
\end{table}

\subsection{Comparing with State-of-the-Art Methods}
\label{sec:sota}
\label{sec:eval-data-annotation-exp}
\parag{Dataset and Test Annotations.} We train on our synthetic data with perfect labels (\cref{sec:synthetic-data-generation}),
and on real data with pseudo labels (\cref{sec:real-data-for-training}).
For a rigorous evaluation, we further require accurate per-point ground truth labels since we cannot rely on the noisy pseudo-labels. As no such dataset exists, we \textit{contribute} new annotations based on EgoBody (please see Sec~\ref{eval-data-annotation}). We define a test split such that there is no overlap of human subjects with the training set. The labeling process is initialized with the noisy pseudo-labels based on the existing multi-view fitted human meshes \cite{egobody}. Expert annotators then manually label the test scenes using an interactive point cloud labeling tool \cite{annotation-tool} to refine the noisy instance masks (illustrated in supplementary material Fig.~\textcolor{red}{6-7}). For body-part labels, pseudo-ground truth labels are refined using the manually corrected instance masks. The test set contains 304 point clouds and 608 humans with various poses, actions, and occlusions. %

\parag{Tasks and Metrics.}
We evaluate our approach on three different 3D point cloud tasks:
human/parts semantic segmentation, human instance segmentation and multi-human body-part segmentation (MHBPS).
For \emph{human} semantic segmentation and \emph{body-part} semantic segmentation,
we report the mean intersection-over-union (denoted as mIoU$^H$ and mIoU$^P$).
For instance segmentation, we use the average precision (AP).
We denote human instance segmentation scores as AP$^H$,
and multi-human body-part segmentation scores (MHBPS) as AP$^P$.
For MHBPS, we additionally report the percentage of correctly parsed body parts (PCP) used by the 2D multi-human parsing community\,\cite{mhpv1}.
Metrics are evaluated at overlaps of $25\%$, $50\%$, and averaged over the range [$0.5$:$0.95$:$0.05$] as in ScanNet\cite{scannet}.

\parag{\name{} Training Details.}
For pre-training and fine-tuning, we train \name{} for 36 epochs each.
We optimize the network with AdamW \cite{Loshchilov19ICLR} and a one-cycle learning rate scheduler \cite{Smith19AIMLMDOA} with a maximal learning rate of $10^{-4}$ and a batch size of $4$ scenes.~Data augmentation includes horizontal flipping, random rotations around the z-axis, elastic distortion \cite{Ronneberger2015MICCAI}, and random scaling by Uniform[$0.9$, $1.1$]. Training (including pre-training and fine-tuning) with 2\,cm voxels takes 5 days on a single NVIDIA RTX 3090 GPU. %

\parag{Methods in Comparison.}
We compare with a wide range of prior-art methods adapted for 3D human segmentation.
MinkowskiUNet\,\cite{minkowski} and KPConv\,\cite{kpconv} are voxel-based and point-based 3D semantic segmentation methods.
Mask3D\cite{mask3d} is the state-of-the-art for 3D instance segmentation.
We additionally compare with two 2D image baselines:
The first one, proposed in \cite{egobody},
obtains human semantic masks from a pre-trained DeepLabv3\cite{deeplab} applied to Kinect RGB images.
Human instance masks come from a pre-trained Mask-RCNN\,\cite{maskrcnn}.
The final 2D human instance masks are the intersection of the semantic and instance masks.
Body-parts are not predicted.
The second baseline, RP\,R-CNN \cite{rprcnn}, is a recent 2D multi-human part segmentation method.
We finetune their checkpoint on our projected 2D EgoBody body-part labels.
For both baselines, we backproject the 2D predictions into 3D for evaluation.

\begin{table}[t]
\centering
\setlength{\tabcolsep}{3pt}
\resizebox{\columnwidth}{!}{
\begin{tabular}{lccccccc}
\toprule  %
      & \multicolumn{3}{c}{\multirow{2}{*}{\em Trained on EgoBody}} && \multicolumn{3}{c}{\em Pre-trained on Synthetic} \\
      & \multicolumn{3}{c}{} && \multicolumn{3}{c}{\em Fine-tuned on EgoBody} \\
       \cmidrule(r){2-4} \cmidrule(r){6-8}
Model & Scene & Human & \cellcolor{gray!10}mIoU$^H$ && Scene & Human & \cellcolor{gray!10}mIoU$^H$\\ %
\midrule
MinkUNet \cite{minkowski}   & $97.5$ & $85.2$ & \cellcolor{gray!10}$91.3$ && $98.0$ & $87.9$ & \cellcolor{gray!10}$92.2$ \textcolor{darkgreen}{\small{(+$0.9$)}} \\
KPConv \cite{kpconv}      & $\mathbf{98.9}$ & $93.4$ & \cellcolor{gray!10}$96.1$ & & $99.1$ & $94.4$ & \cellcolor{gray!10}$96.7$ \textcolor{darkgreen}{\small{(+$0.6$)}}\\
Mask3D \cite{mask3d}       & $98.4$ & $90.9$ & \cellcolor{gray!10}$94.7$ && $99.3$ & $95.9$ &\cellcolor{gray!10}$97.6$ \textcolor{darkgreen}{\small{(+$2.9$)}} \\%
\name{} (Ours)              & $94.5$ & $\mathbf{99.0}$ & \cellcolor{gray!10}$\mathbf{96.8}$ && $\mathbf{99.5}$ & $\mathbf{97.0}$ & \cellcolor{gray!10}$\mathbf{98.3}$ \textcolor{darkgreen}{\small{(+$1.5$)}} \\
\bottomrule
\end{tabular}
}
\vspace{-6px}
\caption{\textbf{3D Semantic Segmentation Scores on EgoBody test.} We perform binary segmentation (scene \vs human). We report per-class (scene \vs human) IoU and mean IoU (mIoU$^H$). For Mask3D and \name{}, human instance masks are merged prior to computing the semantic segmentation scores. Synthetic data pre-training results in improvements of up to +$2.9$\,mIoU$^H$. %
}
\vspace{-1px}
\label{tab:sem_seg_benchmark}
\end{table}

\parag{3D Multi-Human Body-Part Segmentation (MHBPS).}
\reftab{mhp_benchmark} shows MHBPS scores of the baselines and our \name{}.
The task is to detect individual human instance masks and partition them into body parts.
Since there are no existing baseline models that directly predict MHBPS from point clouds,
we construct strong baselines using existing 3D instance \cite{mask3d} and 3D semantic segmentation \cite{kpconv, minkowski} methods and solve two subtasks:
Human instance masks are directly obtained from Mask3D\,\cite{mask3d} or
 by applying density-based clustering HDBSCAN\,\cite{mcinnes2017hdbscan, mcinnes2017accelerated} on the predicted human segments (or body-part segments) from \cite{minkowski, kpconv}.
MHBPS predictions are then obtained by combining human instance masks with semantic segmentation of body parts, \ie{},
predicted body-parts inside a human mask are assigned to that human instance.
Body-parts outside of any human mask are discarded. %

\name{} outperforms all tested combinations of baseline methods including 2D baselines projected to 3D.
Remarkably, \name{} outperforms all prior task-specific methods on 3D instance segmentation (e.g.~Mask3D), and 3D semantic segmentation (e.g.~\mbox{KPConv}) by at least +$3.5$\,AP$^H$ and +$1.6$\,mIoU$^H$. \name{} also significantly improves over the state-of-the-art image baseline RP~R-CNN \cite{rprcnn}
that relies on RGB information and is pre-trained on much larger image datasets.
Notably, we achieve these scores with depth information only.
This demonstrates the benefits of \name{} operating directly on point clouds.

\parag{3D Instance Segmentation.}
Results are shown in Tab.\,\ref{tab:inst_seg_benchmark}.
The task is to predict a set of human instances as binary foreground/background masks over the entire 3D point cloud.
As before, for the 3D semantic segmentation baselines KPConv\,\cite{kpconv} and MinkUNet\,\cite{minkowski}, human instances are obtained by applying density-based clustering HDBSCAN on the predicted human segments (or body-part segments) while Mask3D directly predicts human instance masks.
\name{} largely outperforms all baselines tested, by at least +$3.5$\,AP$^H$.
Moreover, pre-training with synthetic data consistently improves all methods, and is particularly helpful for \name{} (+$8.6$\,AP$^H$) which is key to improved human instance segmentation results.

\begin{figure}[t]
    \centering
    \definecolor{bblue}{HTML}{005FFF} %
\definecolor{bbluel}{HTML}{86ACF8}
\definecolor{rred}{HTML}{D07202} %
\definecolor{rredl}{HTML}{F2A354}

\begin{footnotesize}
\begin{tikzpicture}
    \begin{axis}[
        width  = 0.5*\textwidth,
        height = 0.25*\textwidth,
        major x tick style = transparent,
        ybar=2*\pgflinewidth,
        bar width=10pt,
        ymajorgrids = true,
        ylabel = {{\color{bblue}AP$_{50}^P$}\,\,\,{\color{rred}AP$_{50}^H$}},
        ylabel style={at={(axis description cs:-0.05,0.5)}},
        symbolic x coords={Low Occlusion,Medium Occlusion, High Occlusion},
        xlabel style={at={(axis description cs:-0.5,0.5)}},
        xtick = data,
        scaled y ticks = false,
        enlarge x limits=0.25,
        ymin=67, %
        legend cell align=left,
        legend style={
                draw=none,
                legend columns=2,
                at={(-0.1, -0.58)},
                anchor=south west,
                column sep=0ex
        }
    ]

        \addplot[style={bblue,fill=bbluel,postaction={pattern=crosshatch, pattern color=white},mark=none}] coordinates {(Low Occlusion, 94.6) (Medium Occlusion, 85.1) (High Occlusion, 70.0)}; %

        \addplot[style={bblue,fill=bbluel,mark=none}] coordinates {(Low Occlusion, 98.1) (Medium Occlusion, 94.6) (High Occlusion, 82.1)}; %

        \addplot[style={rred,fill=rredl,postaction={pattern=crosshatch, pattern color=white},mark=none}] coordinates {(Low Occlusion, 99.2) (Medium Occlusion, 94.4) (High Occlusion, 95.1)}; %

        \addplot[style={rred,fill=rredl,mark=none}] coordinates {(Low Occlusion, 100) (Medium Occlusion, 100) (High Occlusion, 100)}; %

        \legend{
        AP${^P_{50}}$ Body\,Part\,{(Real)},
        AP${^P_{50}}$ Body\,Part\,{(Synth+Real)},
        AP${^H_{50}}$ Human\,Inst.\,{(Real)},
        AP${^H_{50}}$ Human\,Inst.\,{(Synth+Real)}}
    \end{axis}
\end{tikzpicture}
\end{footnotesize}
    \vspace{-8px}
    \caption{
    \textbf{Occlusion Analysis.}
    mAP$_{50}$ on EgoBody test on body-part segmentation \colorsquare{bblue} and
    human instance segmentation \colorsquare{rred} for \name{} with and without pre-training on synthetic data.
    Pre-training on synthetic data is particularly helpful for highly occluded humans, \eg{}, part segmentation improves by +12.1\,AP$_{50}^P$.}
    \label{fig:occlusion_analysis}
\end{figure}
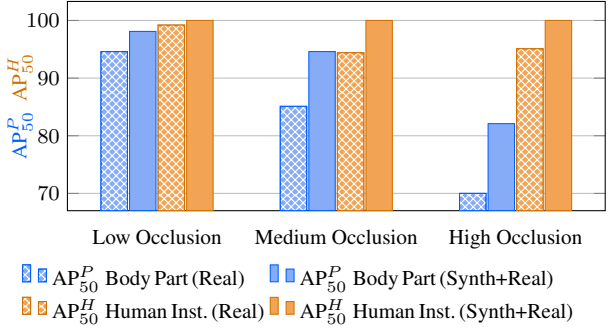

\begin{table}[t]
\vspace{-5px}
\centering
\setlength{\tabcolsep}{3pt}
\resizebox{\columnwidth}{!}{
\begin{tabular}{ll cc c}
\toprule
& & \multicolumn{2}{c}{\em 3D Instance} & {\em 3D Semantic}\\
& & \multicolumn{2}{c}{\em Segmentation} & {\em Segmentation}\\
       \cmidrule(r){3-4} \cmidrule(r){5-5}
{Pre-Training Data} &  {Fine-Tuning Data}
& AP$^H_\text{}$ & AP$^H_{50}$ & mIoU$^H$\\
\midrule
\circlenum{1} -- & Real (EgoBody)  & $89.4$ & $95.4$ & $94.7$ \\ 

\circlenum{2} Real (BEHAVE)  & Real (EgoBody) & $92.0$ & $96.8$  & $96.8$ \\ %
\circlenum{3} Real (EgoBody)  & Real (EgoBody) & $91.8$ & $96.9$ & $95.8$ \\ 
\circlenum{4} Synthetic (ours)  & Real (EgoBody) & $\mathbf{95.6}$ & $\mathbf{98.7}$ & $\mathbf{97.6}$ \\ %
\bottomrule
\end{tabular}
}
\vspace{-8pt}
\caption{\textbf{Training Settings Analysis.}
We compare pre-training on synthetic and real data for instance and semantic segmentation.
}
\label{tab:inst_seg_dataset_analysis_small}
\end{table}

\parag{3D Semantic Segmentation.}
Tab.\,\ref{tab:sem_seg_benchmark} shows binary (scene \vs human) segmentation results with and without pre-training on synthetic data. We adapt Mask3D\cite{mask3d} and \name{} by merging predicted human instance masks with confidence scores above $50\%$ before computing semantic segmentation scores.
We observe that \name{} significantly outperforms specialized semantic segmentation models \cite{minkowski, kpconv} by at least +$1.6$\,mIoU$^H$.
Intuitively, \name{} has the potential to leverage the body-part annotations as an additional supervision signal.
Again, we find that pre-training with synthetic data enhances the performance of all models.

\subsection{Analysis Experiments}
\label{sec:analysis}

\parag{Does synthetic data help with occlusions?}
Occlusions are a main challenge in cluttered indoor spaces.
In Fig.\,\ref{fig:occlusion_analysis}, we analyze the influence of synthetic training data on occluded humans.
One key advantage of synthetic data is that it can be tailored to specific edge cases that are rare in real data.
Our synthetic data contains numerous people in real cluttered scenes and therefore numerous occlusions.
To evaluate the effect of occlusions,
we further split our test data into three groups of increasing levels of human occlusions:
\emph{low} (122 scenes), \emph{medium} (104 scenes), \emph{high} (78 scenes).
Details are in the supplementary.
Pre-training with synthetic data drastically improves body-part segmentation (+$12.1$\,AP$^P_{50}$)
and human instance segmentation (+$4.9$\,AP$^H_{50}$) performance for highly occluded humans.

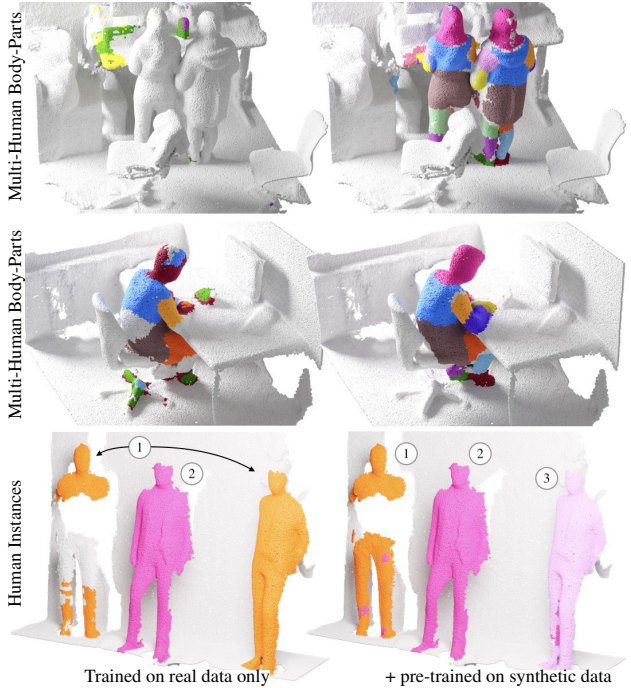
\begin{figure}
\centering
    \begin{overpic}[ width=0.47\textwidth,abs,unit=1mm,scale=.25,trim={0 0 0 1.5cm},clip]{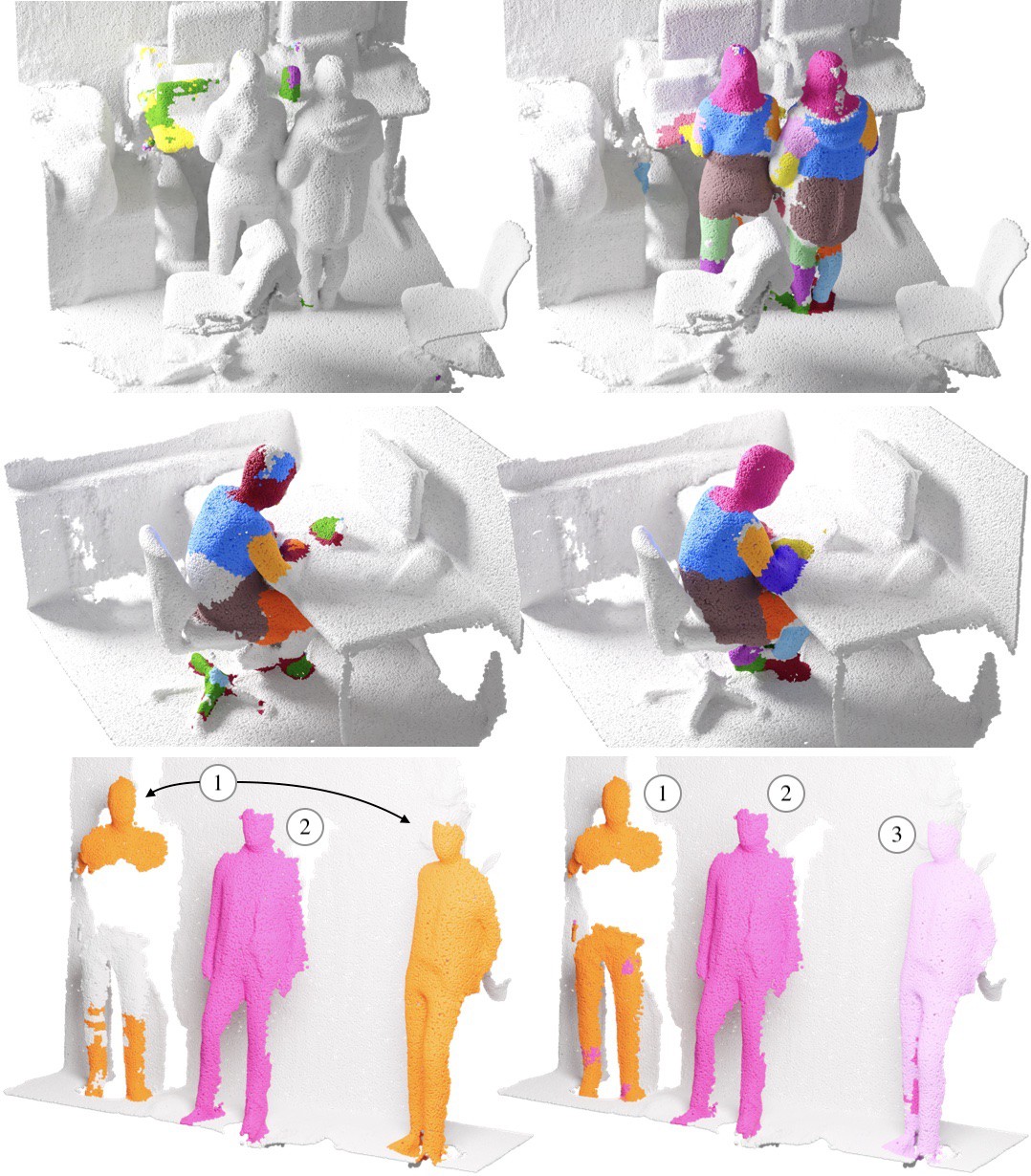}
\put(-0.1,64){\rotatebox{90}{\scriptsize \colorbox{white}{Multi-Human Body-Parts}}}
\put(-0.1,35){\rotatebox{90}{\scriptsize \colorbox{white}{Multi-Human Body-Parts}}}
\put(-0.1,10){\rotatebox{90}{\scriptsize \colorbox{white}{Human Instances}}}
\put(10,0){\scriptsize \colorbox{white}{Trained on real data only}}
\put(50,0){\scriptsize \colorbox{white}{+ pre-trained on synthetic data}}
\end{overpic}
    \vspace{-5px}
    \caption{\textbf{Effect of Synthetic Data.}
    Model trained only on real EgoBody data \emph{(left)} and additionally pre-trained on synthetic data \emph{(right)}.
    Synthetic pre-training improves robustness for close interactions of humans \emph{(top)} or human-scene interactions \emph{(middle)}, and improves generalization to multiple people \emph{(bottom)}.
    }
    \label{fig:synth_real_comparison}
\end{figure}

\begin{figure*}[!t]
  \centering
  \includegraphics[draft=False, width=1.0\linewidth,trim={0 0.3cm 0 0.1cm},clip]{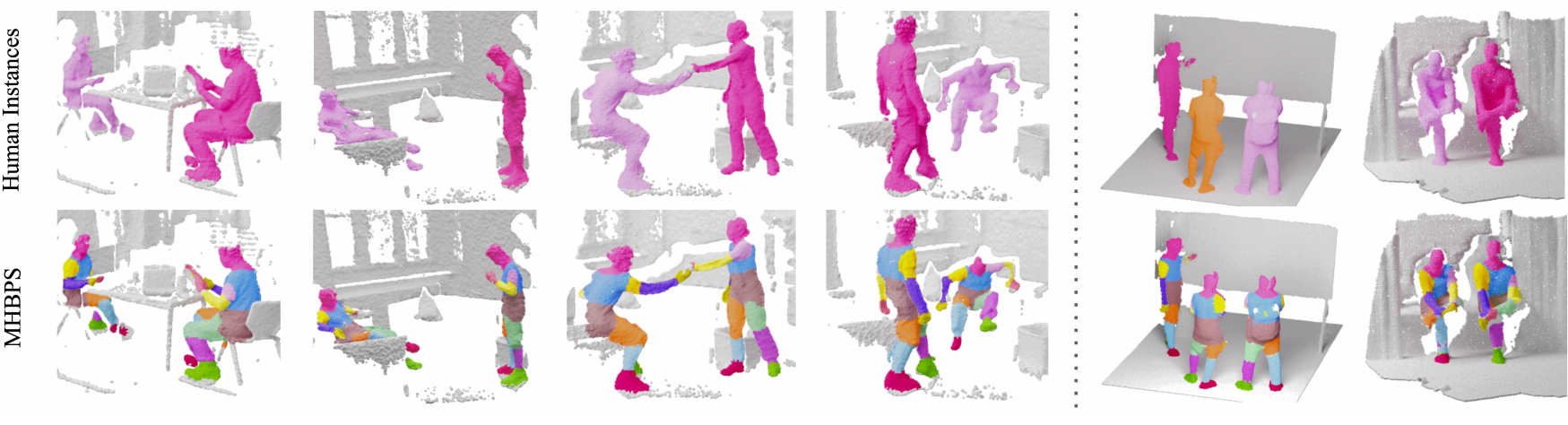}%
{
\vspace{-15px}\\
\resizebox{\textwidth}{!}{
\ColorMapCircle{head}\,Head
\ColorMapCircle{rightArm}\,RightArm
\ColorMapCircle{leftArm}\,LeftArm
\ColorMapCircle{rightForeArm}\,RightForeArm
\ColorMapCircle{leftForeArm}\,LeftForeArm
\ColorMapCircle{rightHand}\,RightHand
\ColorMapCircle{leftHand}\,LeftHand
\ColorMapCircle{torso}\,Torso
\ColorMapCircle{hips}\,Hips
\ColorMapCircle{rightUpLeg}\,RightUpLeg
\ColorMapCircle{leftUpLeg}\,LeftUpLeg
\ColorMapCircle{rightLeg}\,RightLeg
\ColorMapCircle{leftLeg}\,LeftLeg
\ColorMapCircle{rightFoot}\,RightFoot
\ColorMapCircle{leftFoot}\,LeftFoot
}
}
\vspace{-15px}
\caption{
\textbf{\name{} Qualitative Results.}
Human instance segmentation {results} \emph{(top)} and multi-human body-part segmentation results \emph{(bottom)} on point clouds from Kinect sensors from our EgoBody test set \emph{(left)} and on out-of-domain point clouds from iPhone LiDAR scans \emph{(right)}.
The rightmost example shows a failure case where the left and right legs are confused due to the person crossing their legs.
}
\vspace{-10px}
\label{fig:qualitative}
\end{figure*}
\parag{Does synthetic data improve generalization?}
To keep labeling efforts within limits, EgoBody\cite{egobody} does not contain humans that are too closely interacting with other humans or objects,
and is limited to two humans per scene.
A key question is whether synthetic data can help to generalize beyond these limitations of the real-world training scenes.
Fig.\,\ref{fig:synth_real_comparison} depicts these edge cases and shows improved performance when comparing our \name{} with and without pre-training on synthetic data.
The pre-trained model is able to segment humans that are closely interacting \emph{(top)},
a person that is in close contact with a desk and thus heavily occluded \emph{(middle)}, and can successfully segment more than two people
where the model trained on real data assigns the same instance label to two different people \emph{(bottom)}.

\parag{Pre-training on synthetic or real data?}
In a preliminary study (Tab.\,\ref{tab:inst_seg_dataset_analysis_small}), we compare different settings for pre-training on 3D instance and semantic segmentation using \cite{mask3d}.
We always fine-tune on the real EgoBody training set.
The baseline \circlenum{1} does not include any pre-training.
Model \circlenum{4} pre-trained on synthetic data provides the biggest boost
over \circlenum{1} (+$6.2$~AP$^H$, +$2.9$~mIoU).
To verify that the improvement is not due to more training iterations or better weight initialization,
we repeat the experiment and use EgoBody also for pre-training \circlenum{3} as well as another real dataset BEHAVE \circlenum{2}.
We see that \circlenum{2} and \circlenum{3} perform comparably. 
Importantly, however, pre-training on synthetic data \circlenum{4} improves significantly over pre-training on Ego\-Body~\circlenum{3} and BEHAVE~\circlenum{2}, proving the importance of synthetic pre-training.

\parag{\name{} Ablation Study.}
In \reftab{human3d_ablation},
we analyze design choices of \name{}, \ie, the masked attention module, and Hungarian matching.
The study reveals that our newly proposed \emph{two-stage} Hungarian matching is crucial for MHBPS. 
When using the existing \emph{single}-stage Hungarian matching (as in \cite{mask3d, Carion20ECCV}), 
body-part queries and human queries of the same human can be falsely assigned to two different ground truth humans.
Instead, our two-stage Hungarian matching guarantees consistent supervision such that human queries and the corresponding body-part queries are always supervised by a single ground truth human.
The effect of restricting the cross-attention between body-part queries and point features to lie within the corresponding human mask is less significant but improves AP$^P_{50}$ scores.

\begin{table}[!t]
\centering
\setlength{\tabcolsep}{10pt}
\setlength{\tabcolsep}{4pt}
\resizebox{0.47\textwidth}{!}{
\begin{tabular}{cc cccc}
\toprule
Two-stage & Restricted & \multicolumn{4}{c}{\em Multi-Human Body-Part Seg.}\\
\cmidrule(r){3-6}  
Hungarian Matching & Cross-Attention & AP$^P_\text{}$ & AP$^P_{50}$  & PCP & PCP$_{50}$ \\
\midrule
\cmark{}  (two-stage) & \cmark{} & $33.7$ & $\mathbf{82.3}$ & $30.8$ & $66.9$ \\ 
\cmark{}  (two-stage) & \xmark{} & $\mathbf{34.0}$ & $79.5$ & $\mathbf{31.1}$ & $\mathbf{78.1}$ \\
\xmark{} (one-stage)  & \xmark{}& $2.0$ & $12.5$ & $2.2$ & $8.0$\\ 
\bottomrule
\end{tabular}
}
\vspace{-3px}
\caption{\textbf{\name{} Ablation Study.}
Hungarian matching and attention mechanisms. Models trained on EgoBody, no pre-training.
}
\vspace{2px}
\label{tab:human3d_ablation}
\end{table}

\subsection{Qualitative Results and Discussion}
\label{sec:qualitative_results}
\vspace{-3.5px}

Fig.\,\ref{fig:qualitative} shows qualitative results of \name{} for 3D instance segmentation and 3D multi-human body-part segmentation.
Our model works on point clouds from Kinect depth sensors \emph{(left)} and generalizes to out-of-domain point clouds as shown by the scans from the iPhone LiDAR sensor \emph{(right)}.
\name{} is able to {clearly segment} closely interacting humans,
under strong occlusions, and in close contact with scene objects such as sofas or chairs.
This is also reflected in the scores reported in Tab.\,\ref{tab:mhp_benchmark}.
The body-part segmentation can fail when people cross their legs (\ie{}, left/right confusion).
Additional qualitative results are provided in the supplementary material Sec.~\textcolor{red}{5}.

\parag{Limitations.}
Our unified \name{} shows considerable improvements over combinations of specialized state-of-the-art 3D segmentation methods; however, several limitations remain.
Our method focuses on segmenting humans and body parts, while other works \cite{mask3d, minkowski, kpconv} primarily focus on 3D scene segmentation.
In this context, it would be interesting to explore a unified approach that jointly predicts segmentation for both humans and scenes.
Similar to existing work for placing humans into 3D scenes \cite{place, posa,wang2022humanise}, our pipeline generates humans with minimal clothing.
To obtain more realistic training data, a promising avenue would be to integrate the generation of clothed humans \cite{CAPE:CVPR:20, 
 chen2022gdna}.

\vspace{-10px}
\iccvsection{Conclusion}
\vspace{-6px}
\label{sec:discussion}
In this work, we have introduced \name{}, the first unified model for end-to-end 3D multi-human body-part segmentation, operating directly on point clouds.
The key novelties of our transformer-based model are the two-level queries representing human and body-part instances, as well as the two-stage Hungarian matching for supervision. Using our synthetic training data generation framework, we have further shown that pre-training on synthetic training data can significantly improve 3D human segmentation performance on various tasks and models, especially in challenging conditions such as strong occlusion. We believe that \name{} is an important step towards holistic 3D scene understanding with human-scene interactions.

\parag{Acknowledgments.}
This project is funded by Innosuisse (48727.1 IP-ICT), 
ERC CoG grant DeeViSe (ERC-2017-CoG-773161), BMBF project 6GEM (16KISK036K), SNF Grant 200021 204840, and compute resources from RWTH Aachen (rwth1261). We sincerely thank Siwei Zhang for helping with the EgoBody dataset, Anne Marx and Theodora Kontogianni for providing guidance on the 3D annotation tool, and István Sárándi for helpful discussions.
Francis Engelmann is a post-doctoral research fellow at the ETH AI Center.

{\small
\balance
\bibliographystyle{ieee_fullname}
\bibliography{egbib}
}

\nobalance
\newpage
\clearpage

\twocolumn[
\begin{center}
  {\Large \bf \Large{\bf 3D Segmentation of Humans in Point Clouds with Synthetic Data} \\ {\normalfont Supplementary Material} \par}
  \vspace*{12pt}
\end{center}
]

\setcounter{figure}{0}
\setcounter{table}{0}
\setcounter{section}{0}
\setcounter{theorem}{0}

\textit{In this supplementary material, we provide further details about the synthetic data generation framework as well as the label acquisition process for real data. Furthermore, we describe our model architecture and our experimental procedures in more detail. Finally, we present additional quantitative and qualitative results. We will release our code, model and data for research purposes. }

\iccvsection{Synthetic Data Generation Framework}

\begin{figure}[!h]
    \centering
    \includegraphics[width=\linewidth]{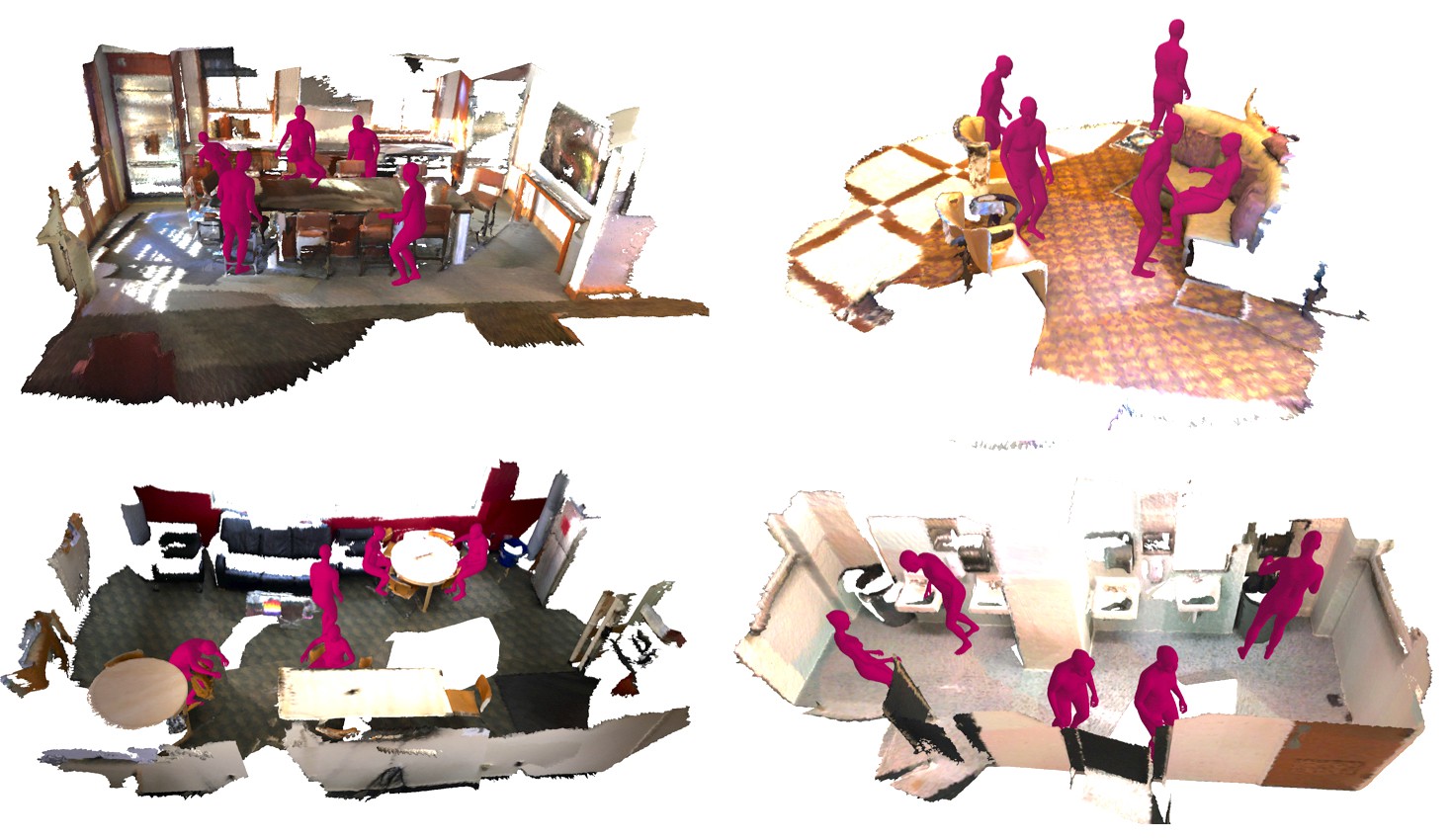}
    \caption{\textbf{Synthetic Humans in ScanNet \cite{scannet} Scenes.} Example scenes (dining room, kitchen, living room, bathroom) populated with synthetic humans using PLACE \cite{place} with instance-segmentation guided human location sampling.%
    }
    \label{fig:scenes_w_humans}
\end{figure}

In this section, we describe our framework for synthesizing virtual humans in realistic environments, and its use for obtaining synthetic training data with perfect ground truth for human instance and body-part segmentation tasks. Our pipeline consists of three main steps: (1) populating 3D indoor scenes (\textit{illustrated in Fig.~\ref{fig:scenes_w_humans}}), (2) rendering depth maps and label images from the 3D indoor scenes with synthetic humans, and (3) obtaining synthetic point clouds with ground truth labels. In the following, we provide details about each component of our pipeline.

\iccvsubsection{Populating 3D Indoor Scenes}
\label{synth-data-framework-supp-populating}
\parag{Real 3D Indoor Scenes.}
In this work, we use 3D real-world scenes from the ScanNet dataset \cite{scannet}, which is a large-scale 3D indoor dataset.
The ScanNet \cite{scannet} dataset features $1513$ scenes and $707$ rooms, and provides 3D surface reconstructions, 3D camera poses, captured RGB-D sequences, as well as annotations for segmentation tasks.
We extend the ScanNet \cite{scannet} dataset by generating synthetic humans in realistic poses, interacting with scenes from the dataset. Please note that there are several other available 3D indoor datasets such as \cite{s3dis, matterport, rio}, and our pipeline can be easily adapted to these datasets as well.

\parag{Scene Boundaries.} 
The synthetic human generation method on which we base our approach, PLACE \cite{place}, requires the computation of scene boundaries as well as the signed distance field (SDF) for each input scene.  Therefore, we first compute the SDF and scene boundaries for all training scenes in ScanNet \cite{scannet}.
The SDF value is $0$ on the surfaces or boundaries of a set, which is utilized by PLACE to find suitable surfaces to place synthetic humans.

\paragraph{PLACE: Proximity Learning of Articulation and Contact in 3D Environments \cite{place}.}
For placing synthetic humans, we leverage PLACE \cite{place}, which is a generative human-scene interaction synthesis method. Given a 3D scene without humans, generation and placement of synthetic humans using PLACE \cite{place} consists of several stages. First, a 3D cage within the scene is randomly sampled, and is transformed into the unit sphere for the computation of Basis Point Set (BPS) encoding of the scene, as well as the scene features. Then, a conditional variational autoencoder (cVAE) is utilized to generate body features conditioned on the scene features of the given 3D cage. Based on the scene BPS and the body features, a regressor is used to predict a set of body mesh vertices, which are then transformed back to the original world coordinate system. In PLACE \cite{place}, the size of the 3D cage is chosen such that the cage is large enough to contain the full body mesh as well as the \textit{nearby} scene objects. The 3D cage size is set as $2 m^3$ following these constraints. Please see PLACE \cite{place} for details. %

\parag{Modified PLACE: Instance Segmentation Guided Human Location Sampling.} PLACE \cite{place} does not give full control over the interaction objects, which poses a limitation for our application as we are primarily interested in capturing humans in various poses with close human-scene interactions.
Hence, we modify the PLACE \cite{place} pipeline to address the need for selecting potential interaction objects and sample potential human locations, guided by object instance labels.
In our modified pipeline, we use ground truth object instance labels from the ScanNet \cite{scannet} dataset to identify areas in which the synthetic humans can closely interact with the scene.
We identify the following object classes as suitable for our use case: \textit{chair, couch, coffee table, seat, bed, table, bench, kitchen counter, sofa, dining table}. 

The Human location sampling process in our modified pipeline consists of the following steps:

(1)~For a given ScanNet scene, we first uniformly sample the number of humans, $n_\text{humans} \in [5, 10]$. 

(2)~Using the ground truth instance labels, we then identify $n_\text{objects}$, the number of object instances from the selected object categories present in the given scene. 

(3)~If $n_\text{objects} >= n_\text{humans}$, we uniformly sample a subset of the object instances to select $n_\text{humans}$ objects. Otherwise, we select all available ($n_\text{objects}$) objects, and then randomly sample $n_\text{random\_cage}=n_\text{humans}-n_\text{objects}$ following the original implementation to reach the intended number of bounding boxes to place humans. We use the same 3D cage size ($2 m^3$) as used for training PLACE \cite{place}. 

(4)~Using the selected bounding boxes, we follow the BPS encoding, scene feature extraction and human body synthesis stages from PLACE. We use $200$ and $100$ iterations for the simple and advanced optimization of PLACE, respectively. Moreover, we increase the weight of the collision loss term (from $8.0$ to $10.0$) in the advanced optimization to reduce inter-penetrations.

Overall, our pipeline enables us to generate humans in various poses while taking human-scene proximity into account for close interaction scenarios (with scene objects such as \textit{tables} and \textit{chairs}).

\iccvsubsection{Rendering} We are primarily interested in creating a labeled synthetic dataset of partial point clouds obtained from depth scans. In order to obtain realistic depth maps and corresponding label images, we need to place a virtual camera in each scene with synthetic humans, and render frames using this virtual camera. With this purpose, we employ a simple virtual camera placement procedure. %

First, we compute the scene center as the arithmetic mean of the global vertex coordinates of the full scene mesh. In order to better reflect the camera-to-ground distance of a potential handheld capture device (\eg mobile phone, tablet), we uniformly sample a height value from the range $h_c \in [1.4, 1.6]$ m. We place the camera center at the scene center, and then apply a translation to ensure its z-coordinate is equal to the sampled height value $h_c$. Essentially, the camera is always aligned with the ground-plane, i.e., parallel to the xy-plane, however its height and viewing direction may change. We define the viewing direction as the rotation around the z-axis, and uniformly sample this rotation value within $[0^\circ, 360^\circ)$.

For any given scene with synthetic humans, we sample $40$ frames -- please note that one can arbitrarily increase the number of frames captured from a given 3D scene, and easily increase the scale of the dataset. At each rendering iteration, we re-sample the camera-to-ground distance and camera viewing direction. We render depth maps and label images with a resolution of $480 \times 640$ $(h \times w)$ with $60$ degrees of horizontal FOV to imitate a Kinect depth sensor.  %

\begin{figure}[!h]
    \centering
    \includegraphics[width=0.95\linewidth,]{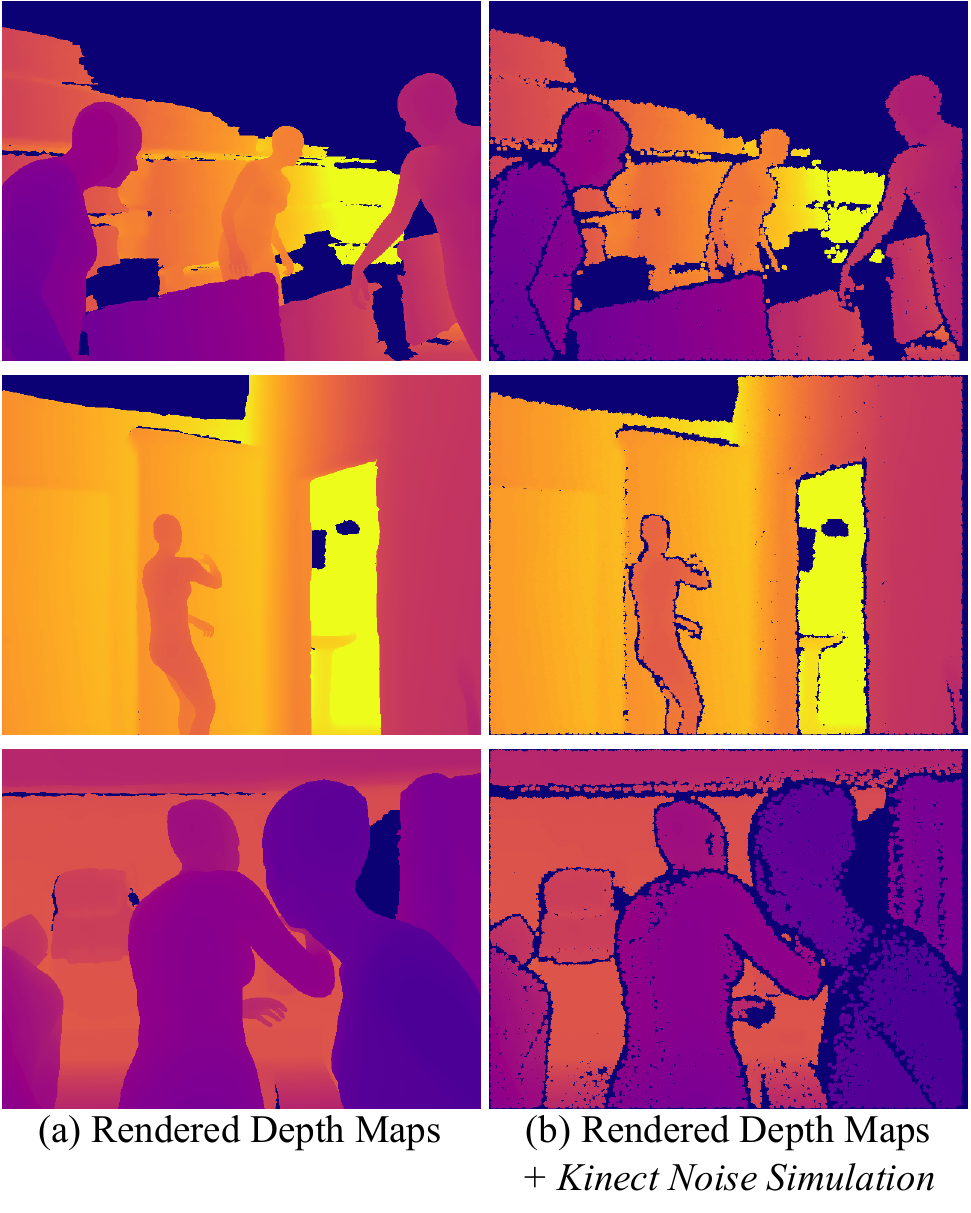}%
    \vspace{-4mm}
    \caption{\textbf{Kinect depth sensor noise simulation.} (a) Using the described rendering pipeline, depth maps are rendered from scenes populated with synthetic humans, (b) simulated Kinect depth sensor noise is applied to the rendered depth maps.}
    \label{fig:kinect_depth_simulation}
\end{figure}

\iccvsubsection{Kinect Depth Sensor Noise Simulation} In order to simulate Kinect depth sensor noise, we use SimKinect \cite{simkinect} -- particularly the implementation available at \cite{simkinect-github}. For each depth image, we perform the noise simulation procedure using a scale factor of $100$, baseline of $0.075$\,m, standard deviation of $0.5$, filter size of $6$, near-plane depth of $0.01$\,m and far-plane depth of $20$\,m. Noise simulation examples are shown in Fig.~\ref{fig:kinect_depth_simulation}.

\begin{figure}[!ht]
    {
    \centering
    \includegraphics[width=0.95\linewidth, trim={0 0.5cm 0 1.8cm},clip ]{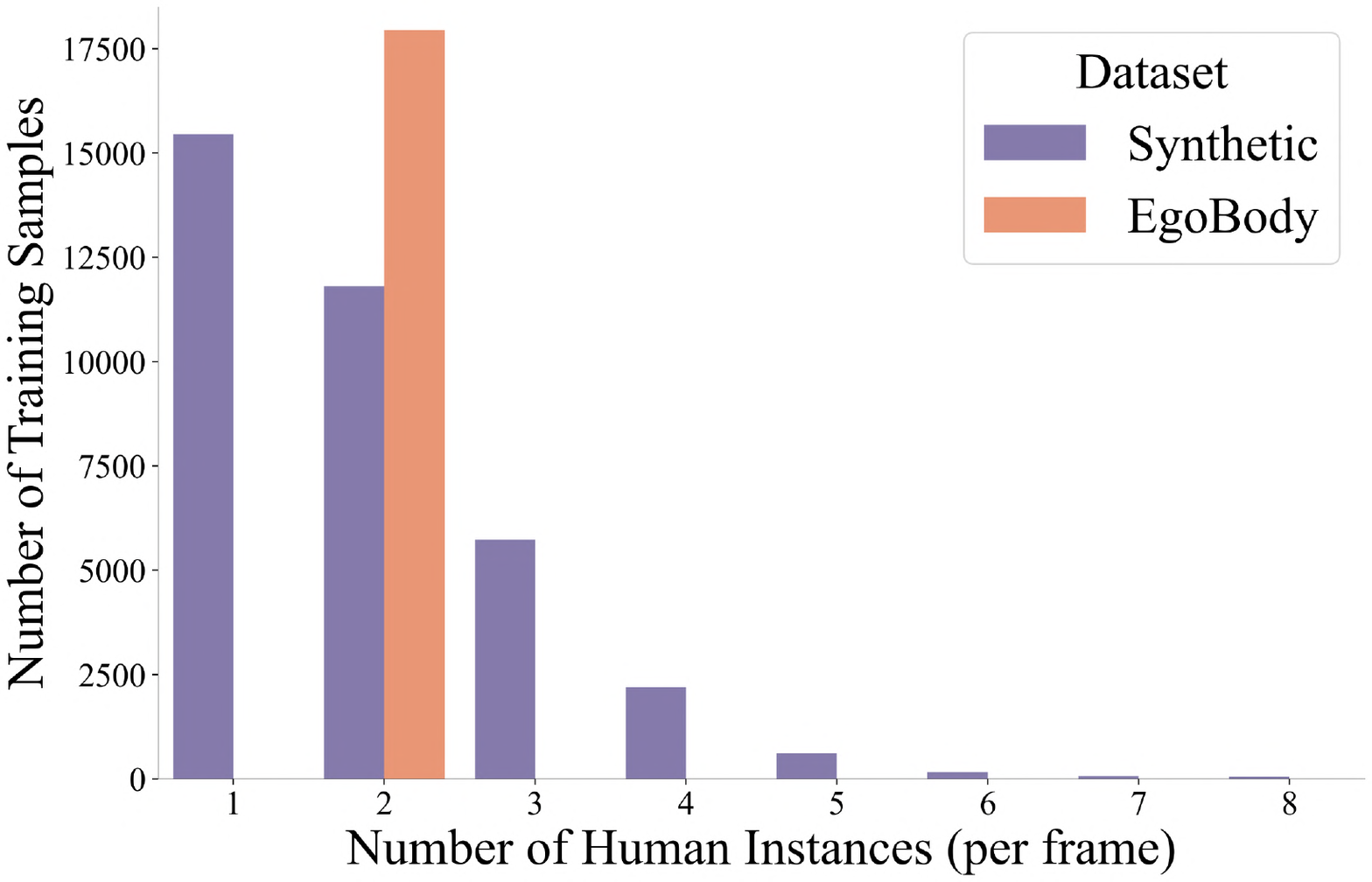}%
    \vspace{-0.5cm}
    \caption{\textbf{Number of human instances vs. number of training samples.}
    Our synthetic dataset features scenes with up to 8 human instances whereas each EgoBody scene features exactly 2 subjects. %
    }
    \label{fig:dataset_stats}
    \vspace{-0.1cm}

    \centering
    \includegraphics[width=\linewidth]{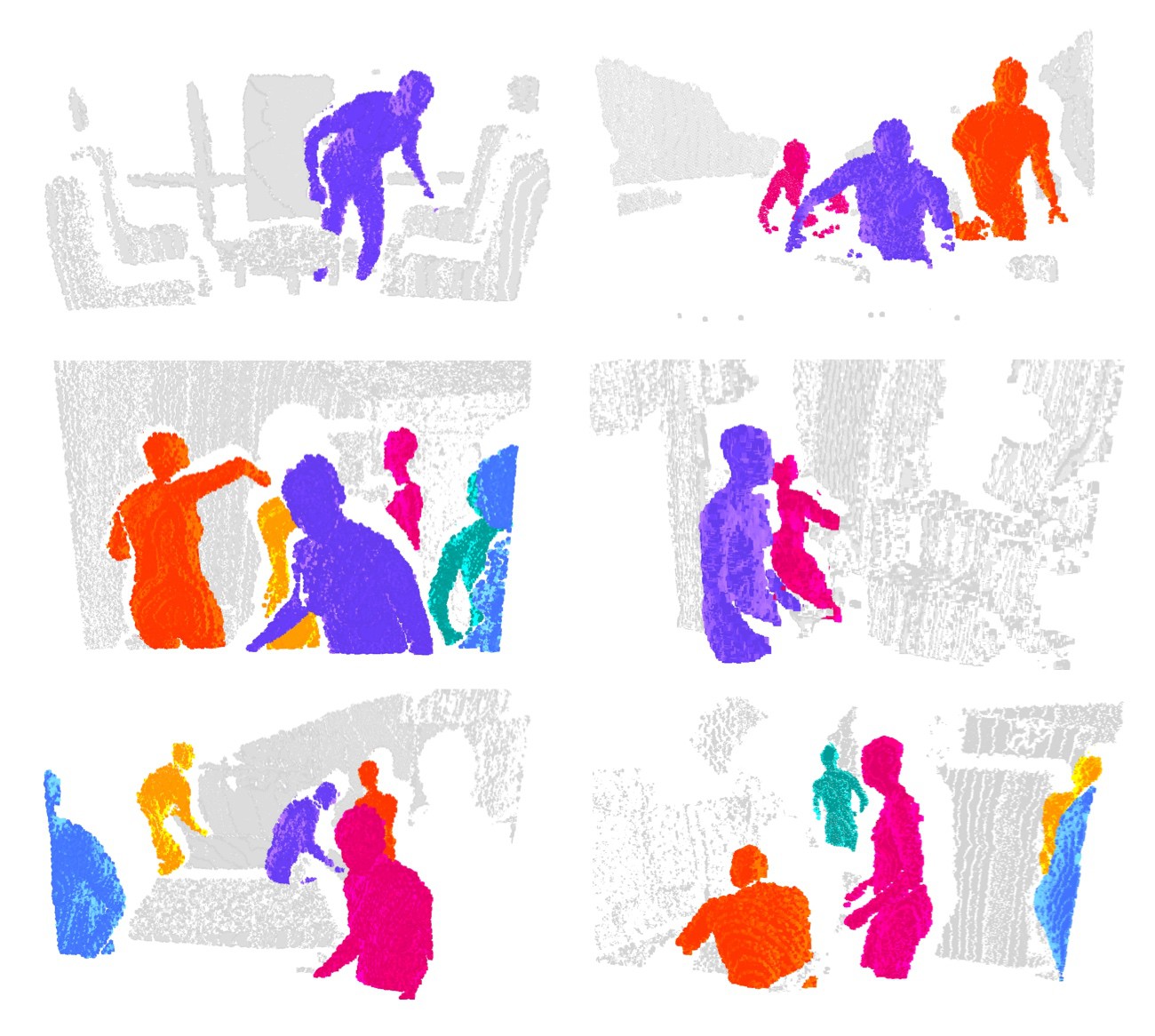}
    \caption{\textbf{Example synthetic training scenes (point clouds).} Our synthetic dataset features point clouds with a varying number of human instances.
    }
    \label{fig:showcase_instances}
    }
\end{figure}

 \iccvsubsection{How many humans are there in each scene?} As described earlier in Sec.~\ref{synth-data-framework-supp-populating}, the number of humans is uniformly sampled in $[5, 10]$ for each of the $1201$ training and $312$ validation scenes from the ScanNet \cite{scannet} dataset. Since the rendering process captures only a portion of the 3D scene based on the sampled camera pose, the number of humans in each \textit{rendered view} in the synthetic dataset is often smaller, and it varies in $[0, 8]$. In contrast, the EgoBody dataset \cite{egobody} only has 2 humans per scene, and the BEHAVE \cite{behave} dataset only features 1 subject per scene. Please see Fig.~\ref{fig:dataset_stats} for an illustration of the number of human instances (per frame) vs. number of training samples.

In Fig.~\ref{fig:showcase_instances}, we show example point clouds (obtained by back-projecting rendered depth maps) from our synthetic dataset, illustrating the varying number of human instances.

\begin{figure}[!pht]
    \centering
    \includegraphics[width=0.90\linewidth]{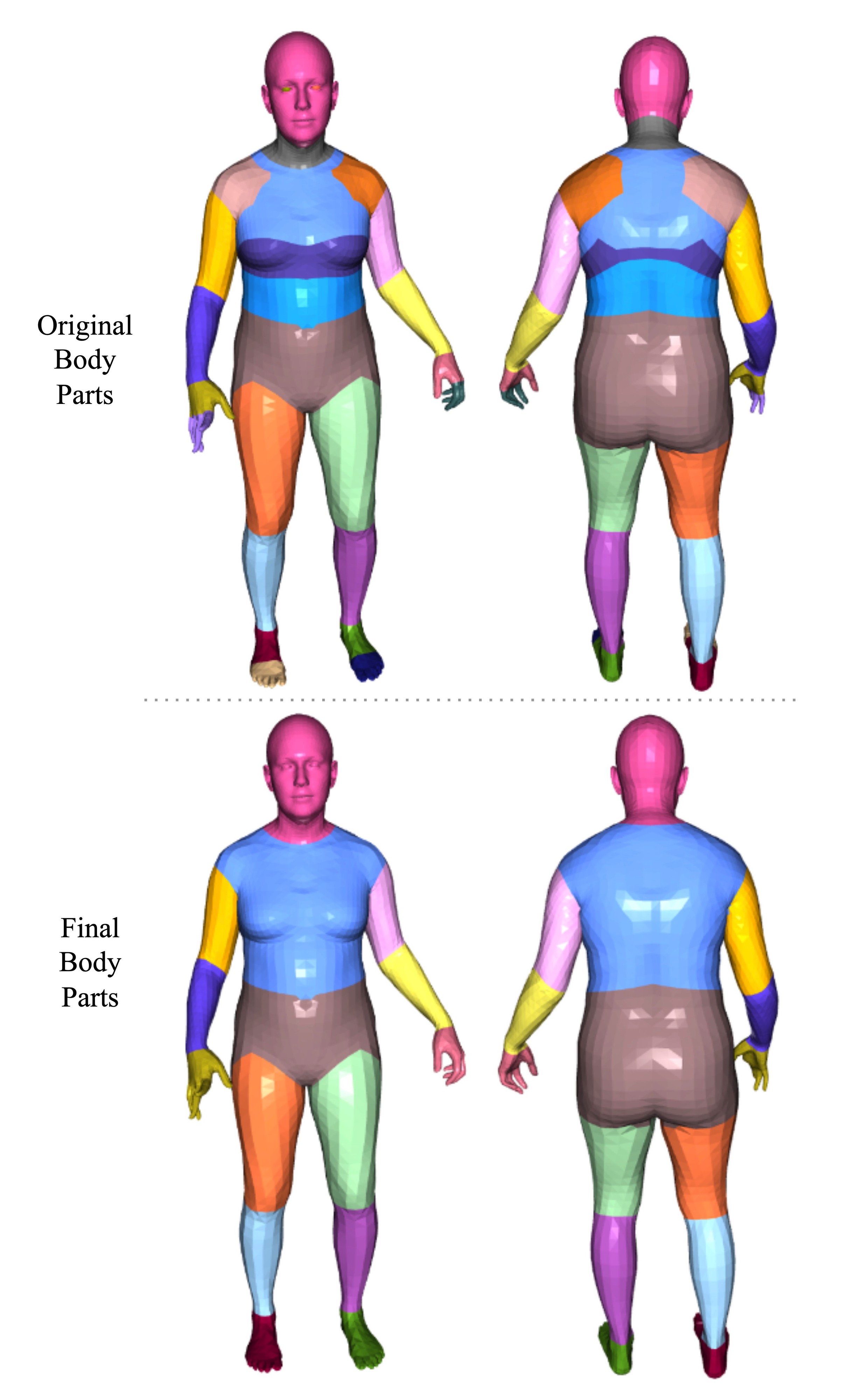}
    \caption{\textbf{Illustration of body part merging.}
    The first row shows original body parts, and the second row shows the body parts obtained after merging smaller parts into larger ones.
    }
    \label{fig:merging}
\end{figure}

\iccvsubsection{Merging Body Parts}
\label{sec:merging_body_parts}
In order to obtain body-part labels, we first map the faces of each SMPL-X \cite{smplx} mesh to 26 body parts according to the mapping in \cite{meshcapade}. Afterwards, we merge smaller body parts into larger ones as shown in Tab.~\ref{tab:merging} and Fig.~\ref{fig:merging}, and obtain 15 body part classes. We follow this merging scheme for all of our experiments (training and evaluation).

\begin{table}[!ht]
\centering
\setlength{\tabcolsep}{3pt}
\resizebox{\columnwidth}{!}{
\begin{tabular}{ll}
\toprule
Merged Body Parts & \cellcolor{gray!10}Final Body Part \\
\midrule
leftEye, rightEye, neck, head & \cellcolor{gray!10}head \\
leftToeBase, leftFoot & \cellcolor{gray!10}leftFoot \\ 
rightToeBase, rightFoot & \cellcolor{gray!10}rightFoot \\ 
leftHandIndex1, leftHand & \cellcolor{gray!10}leftHand\\
rightHandIndex1, rightHand & \cellcolor{gray!10}rightHand\\
spine, spine1, spine2, leftShoulder, rightShoulder & \cellcolor{gray!10}torso\\ 
\bottomrule
\end{tabular}
} 
\vspace{-6px}
\caption{
\textbf{Merged Body Parts.} Smaller body parts (\eg eyes) were merged into larger ones (\eg head)
}
\label{tab:merging}

\end{table}

\begin{figure*}[!t]
  \centering
   \includegraphics[draft=False, width=0.98\linewidth]{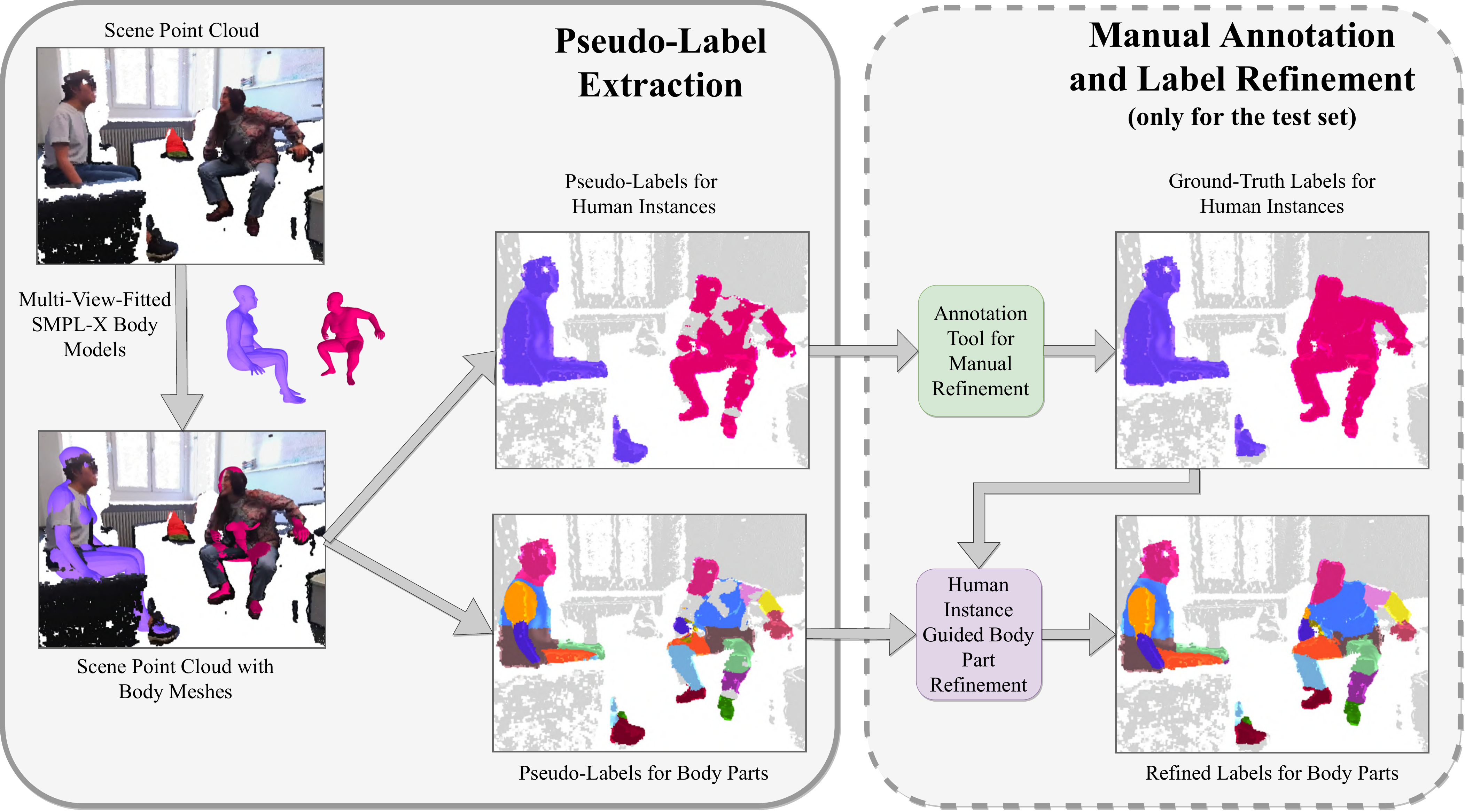}%
 \vspace{-5px}
\caption{
\textbf{Pseudo-label extraction and label refinement.} Pseudo-labels for human instances and body parts are obtained by performing the following procedure for each scene in EgoBody and BEHAVE: Each point in the point cloud obtained from a depth image is assigned to a human instance mask and a body part based on the distance between each point and its closest neighbor in the fitted body mesh. Only for the test set (EgoBody), expert annotators manually refine the human instance masks, which are then used to refine the body part labels.}

\vspace{-10px}

\label{fig:pseudo_label_pipeline}
\end{figure*}

\iccvsubsection{Obtaining Labeled Synthetic Point Clouds} Rendered depth maps are backprojected to the 3D space, along with the label images to obtain perfectly labeled point clouds. After leveraging the depth images with simulated Kinect noise to backproject our label maps, we obtain partial point clouds which can occasionally be very sparse due to the virtual camera viewing direction as well as the simulated noise. Therefore, we perform a post-processing step to remove the scenes with less than $20$k points.  We use this pipeline to create a synthetic dataset for human semantic, human instance, and multi-human body-part segmentation tasks. For semantic and instance segmentation, we provide two labels: background and human. For multi-human body-part segmentation, we map the faces of each SMPL-X \cite{smplx} mesh to body-parts according to the mapping described in Sec.~\ref{sec:merging_body_parts} and assign each point to one of the $15$ body parts.

\iccvsubsection{Dataset Size and Statistics}
We place humans in $1201$ training and $312$ validation scenes from ScanNet, and render (capture) $40$ frames per scene. Samples with fewer than $20$k points are filtered out. Our final synthetic dataset consists of $36536$ training and $12165$ validation samples. For comparison, Real~(EgoBody) dataset has $17943$, and Real~(BEHAVE) dataset has $41088$ training samples. %

\iccvsection{Real Data Collection} %
In this section, we share details about our real data collection, processing and annotation pipelines. %

\begin{figure*}[!t]
  \centering
   \includegraphics[draft=False, width=1.0\linewidth]{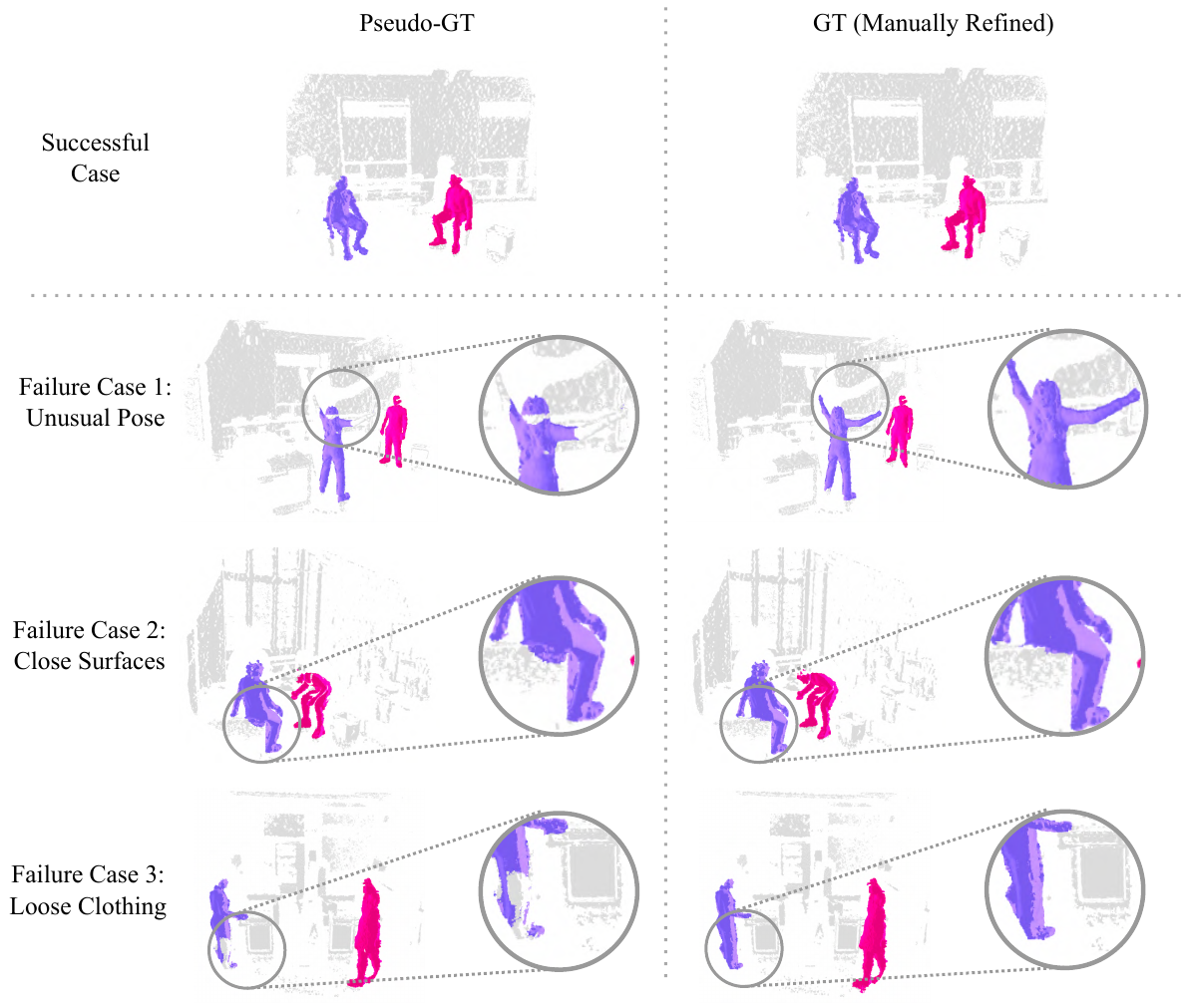}%
\vspace{-15px}
\caption{
\textbf{Pseudo-Ground Truth \vs Manually-Refined Ground Truth}. Pseudo labels may fail particularly in the presence of (1) unusual poses, (2) nearby object- or scene-surfaces, and (3) loose clothing.
Our manual annotations for human instances correct these failure cases (highlighted with circles),
and provide an accurate and reliable evaluation benchmark. 
\vspace{-7pt}
}
\label{fig:pseudo_vs_manual}
\end{figure*}

\iccvsubsection{Pseudo Training Labels on Real Data}
\label{real-data-for-training}
In this section, we give further details about our process for extracting pseudo ground truth labels for human semantic segmentation, instance segmentation and body part segmentation for the EgoBody \cite{egobody} and BEHAVE \cite{behave} datasets. Our pipeline for extracting pseudo-labels is illustrated in Fig.~\ref{fig:pseudo_label_pipeline} (left block).

\parag{EgoBody \cite{egobody}.}
Each EgoBody scene features two subjects captured from multiple Kinect RGB-D cameras (3 or 5 cameras depending on the interaction sequence). Multi-view fitted SMPL-X \cite{smplx} body parameters per each human are available.
We process the frames at 1 FPS. We obtain the human instance masks by selecting scene points under $5$ cm distance to the fitted body mesh. In order to obtain body-part segmentation labels, we first map the faces of each SMPL-X \cite{smplx} body mesh to body parts according to the mapping in \cite{meshcapade}, and merge smaller body parts into larger ones. Then we assign each point in the human mask to the body part category of its closest neighbor in the fitted SMPL-X body mesh.

\parag{BEHAVE \cite{behave}.} In each scene, there is one subject interacting with one object in a largely empty scene captured from 4 Kinect RGB-D cameras. Multi-view fitted SMPL \cite{smpl} body model parameters are available. We obtain the human instance mask by selecting scene points under $5$ cm distance to the fitted SMPL body mesh. As human point clouds were also released with the BEHAVE \cite{behave} dataset, we use these masks to refine the human instance masks we compute based on the distance between each point and its closest neighbor in the fitted body. In order to obtain body-part segmentation labels, we first map the faces of each SMPL \cite{smpl} mesh to body parts according to the mapping in \cite{meshcapade}, resulting in 24 body parts (fewer than SMPL-X, where left-eye and right-eye are also specified as separate body parts), and merge the body parts (see Sec.~\ref{sec:merging_body_parts}). Then we assign each point in the human mask to the body part category of its closest neighbor in the fitted body mesh.

\iccvsubsection{Manually Refined Evaluation Dataset}
\label{eval-data-annotation-supp}
The EgoBody\,\cite{egobody} dataset contains $125$ interaction sequences captured by $3$ or $5$ Kinect cameras depending on the sequence. As the originally published train/validation/test splits were created based on separating first-person view subjects (the subject observed by the other subject wearing a head-mounted device) in each sequence, we created a new split such that none of the subjects overlap across splits.
The split consists of 73 training sequences, 11 validation sequences, as well as 38 test sequences, while 3 sequences were removed to maintain a non-overlapping distribution of subjects across splits.  From each of the test sequences, expert annotators have annotated 8 scenes, resulting in a test set consisting of 304 point clouds featuring a large variety of human poses, action types and occlusion levels. There is potential to expand the test set with a larger number of annotated test scenes in the future.

The annotation was performed using a 3D annotation tool \cite{annotation-tool}. The annotation tool is initialized with pseudo-labels for human instances. Then, the human instance masks are manually refined by annotators, as illustrated in Fig.~\ref{fig:pseudo_label_pipeline} (right block, dotted line). Body part label refinement is guided by the resulting ground-truth human instance masks such that each point in the human mask is assigned to the closest body part in the original fitted body (please see Sec.~\ref{real-data-for-training}), and each point outside of the refined human mask is removed from the body part mask.  %

\iccvsubsection{Pseudo \vs{} Manually Refined Labels}

Although the pseudo-ground truth labels for human masks and body parts were extracted using multi-view fitted body models from EgoBody, the labels can be noisy or incorrect in certain scenarios. Therefore, to obtain a more reliable evaluation set to conduct a thorough evaluation, we refined the instance segmentation masks initialized by the fitted SMPL-X body meshes, following the annotation procedure described in Sec.~\ref{eval-data-annotation-supp}. In \reffig{pseudo_vs_manual}, we illustrate the need for manual refinement, especially in case of close-contact interactions with scene objects (\eg sitting on a sofa), loose clothing (\eg wide-legged jeans) or unusual poses (causing a mismatch between the fitted body mesh and real human point cloud). Furthermore, we quantified the quality of the pseudo-ground truth labels by computing AP scores between the pseudo labels and the manually refined ground truth labels, resulting in $\mathrm{AP}^H:\,91.9$, $\mathrm{AP}^H_{50}:\,99.3$, and $\mathrm{AP}^H_{25}:\,99.5$, highlighting the need for manual annotations.%

\vspace{-3pt}
\iccvsection{\name{} Architecture Details}
\vspace{-2pt}
We obtain strong multi-scale point features from a Minkowski Res16UNet18B\,\cite{minkowski}.
We extract all 5 feature maps of sizes (256, 128, 128, 128, 128) from the U-Net decoder, pass them through a non-shared linear layer in order to project these point features to the transformer decoder features with 128 channels.
Following Mask3D\,\cite{mask3d}, we also use the modified transformer decoder of Mask2Former\,\cite{Cheng22CVPR} instantiated with 8-headed attention and a feedforward network of $1024$ dimensions.
We sample point features for the cross-attention following Mask3D\,\cite{mask3d}.
\name{} learns parametric human and body-part queries during training time.
We assign $16$ body-part queries to each of the $5$ human queries.
Following\,\cite{mask3d,Misra21ICCV}, we use Fourier positional encodings based on normalized voxel positions.
The full model, \ie feature backbone and transformer decoder, uses 18.9 million parameters.

\iccvsection{Experiments}
In this section, we share further details about our experiments presented in the main paper, and provide additional results.
\iccvsubsection{Clustering Details}
For the semantic segmentation baselines KPConv \cite{kpconv} and MinkUNet \cite{minkowski}, we obtain human instances by applying density-based clustering HDBSCAN \cite{mcinnes2017hdbscan, mcinnes2017accelerated} on the predicted human segments or body-part segments. We conduct a hyperparameter study to tune the parameters of the HDBSCAN algorithm, then we set HDBSCAN's minimum number of samples to $1200$, and minimum cluster size to $1500$. Each detected cluster of HDBSCAN represents a spatially contiguous instance. We assign each instance a confidence score of $100\%$.

\iccvsubsection{Performance for Different Activity Types}
We conduct an analysis to assess the effect of pre-training with synthetic data with respect to different human activities. With this purpose, we create a set of activity categories as shown in Tab.~\ref{tab:activity_types}, and manually annotate activities in each test scene. Please note that due to the nature of the dataset, our activity splits partly overlap. There are two main reasons for this. First, each EgoBody scene consists of two human subjects who potentially participate in different types of activities. In such cases, we assign the scene to both activity groups. Second, if subjects take part in compound activities (e.g. sitting down while pointing at an object on the table), we assign the scene to all relevant activity groups.

For each activity group we create, we report average precision scores for body parts (AP$^P_{50}$) with and without synthetic pre-training in Tab.~\ref{tab:activity_types}. While pre-training with synthetic data results in consistent improvements on each activity category, we observe the largest improvement for actions that cause significant self-occlusions such as bending or walking. 

\begin{table}[!ht]
\centering
\setlength{\tabcolsep}{4pt}
\resizebox{\columnwidth}{!}{
\begin{tabular}{l |c c c c c c c c c}
\toprule
\vspace{-12px}
&&&&&&&&& \\[0em]
             &        &        &        & sit down, & lean,    & dance,   & kneel,  & pick, put,     & reach, touch, \\
Human3D      & sit    & stand  & walk   & stand up & lie down & exercise & bend   & hold an object & point at      \\
\midrule
w/o synth. & $84.0$              & $87.9$              & $74.1$              & $86.6$                & $80.7$                & $89.6$                & $81.3$              & $85.2$         & $85.6$        \\
w/ synth.  
& $\mathbf{90.9}$
& $\mathbf{94.0}$ 
& $\mathbf{90.0}$ 
& $\mathbf{92.7}$ 
& $\mathbf{87.1}$%
& $\mathbf{98.2}$%
& $\mathbf{92.1}$%
& $\mathbf{90.1}$%
& $\mathbf{90.0}$%
\\
& \textcolor{ourGreen}{\scriptsize+$6.9$} 
& \textcolor{ourGreen}{\scriptsize+$6.1$} 
& \textcolor{ourGreen}{\scriptsize+$15.9$} 
& \textcolor{ourGreen}{\scriptsize+$6.1$} 
& \textcolor{ourGreen}{\scriptsize+$6.4$}
& \textcolor{ourGreen}{\scriptsize+$8.6$}
& \textcolor{ourGreen}{\scriptsize+$10.8$}
& \textcolor{ourGreen}{\scriptsize+$4.9$}
& \textcolor{ourGreen}{\scriptsize+$4.4$} \\

\bottomrule
\end{tabular}
}
    \caption{\textbf{Multi-Human Body-Part Segmentation Performance for Different Activity Types.} For each activity group, we report average precision scores for body parts (AP$^P_{50}$) with and without synthetic pre-training. We observe the largest improvement for actions that cause significant self-occlusions such as walking and bending. }
    \label{tab:activity_types}
\end{table}

\begin{figure*}[!ht]
  \centering
   \includegraphics[draft=False, width=0.95\linewidth]{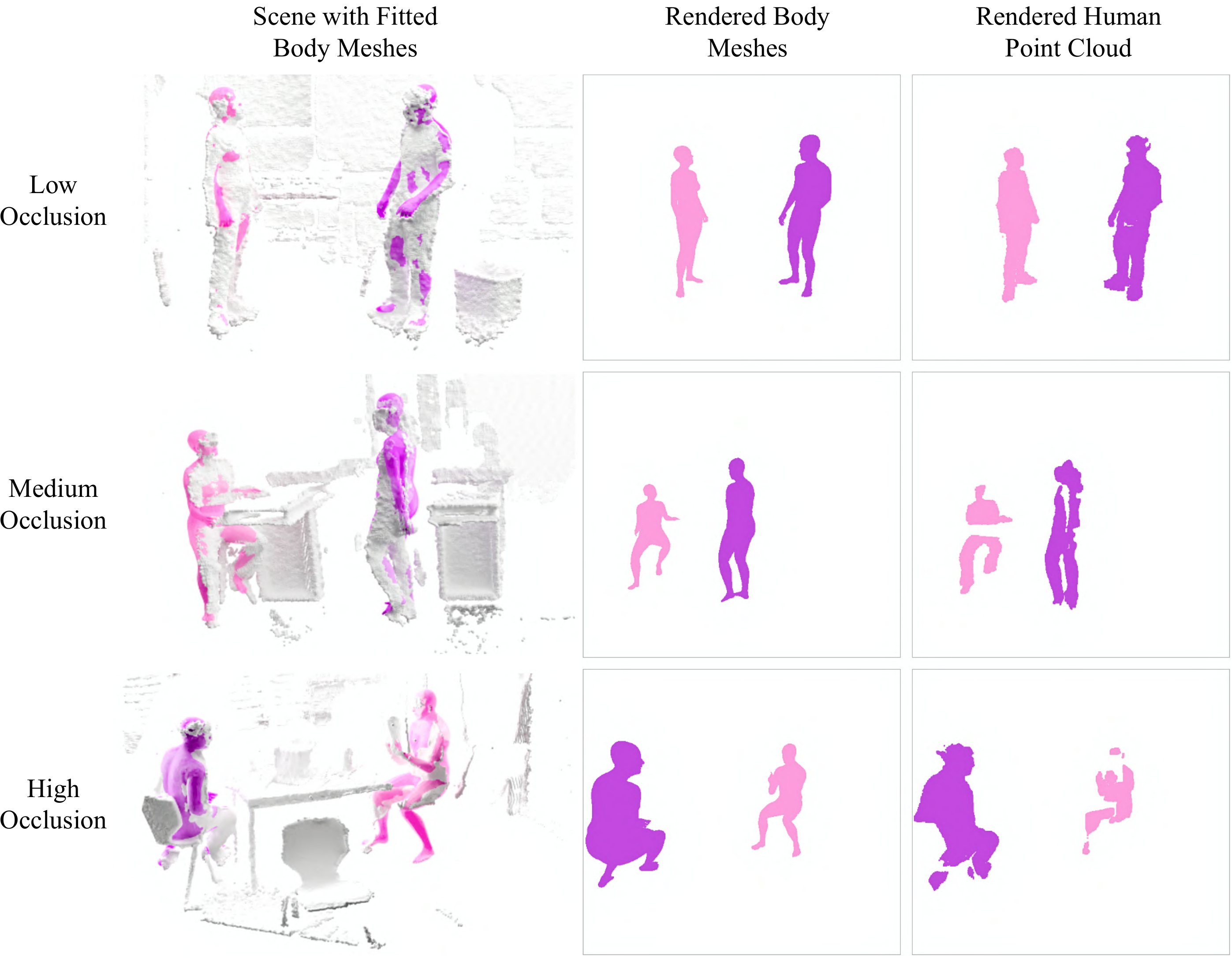}
 \vspace{-5px}
\caption{
\textbf{Occlusion Computation.} In the first column, SMPL-X human body meshes fitted to humans in the EgoBody dataset are shown. The fitted body meshes for each human (\textit{second} column) as well as human masks (\textit{third} column) obtained from our manual annotation of the scene point clouds are projected onto an image. Using these rendered images, it is possible to compute an approximation of the occlusion level. }

\vspace{0cm}

\label{fig:occlusion_computation}
\end{figure*}

\iccvsubsection{Occlusion Computation}
\vspace{-3px}
In the main paper, we have shared our results from an analysis we conducted in order to assess the robustness of our model to occlusions. With this purpose, we split our test dataset into three groups based on the level of human occlusions: low (122 scenes), medium (104 scenes), high (78 scenes). For each scene in the EgoBody test set, we approximate the occlusion level of each human. To this end, we first project the fitted SMPL-X human body meshes for each human onto an image (see Fig.~\ref{fig:occlusion_computation}, \textit{second} column). Then, we project the human masks obtained from our manual annotation of the scene point cloud (see Fig.~\ref{fig:occlusion_computation}, \textit{third} column). Using these rendered images, it is possible to compute an approximation of the occlusion level. The occlusion level is inversely proportional to the ratio between the pixel-wise area of the human mask, and the area of the rendered body mesh. The computed ratio is only an \textit{approximation} of the actual visibility, as the fitted body meshes are not perfect, and points are sometimes sparse in certain parts of the body due to Kinect depth noise. Each test scene consists of two human subjects, and we classify each scene based on the occlusion level of the highest occluded subject. Using this procedure, we first obtained an initial grouping based on the approximated visibility, which was then followed by a manual iteration to correct and account for potential mismatches between the fitted body and actual human mask. %

\iccvsubsection{Comparison to Image Baseline}

Our approach is the first human segmentation method to operate directly on 3D point clouds of cluttered scenes.
In the main paper, we compared our approach with two image-based baselines that operate on color images and project the segmentation masks onto the 3D point cloud obtained from the Kinect depth map.

\begin{figure}[t]
    \centering
    \includegraphics[width=\linewidth]{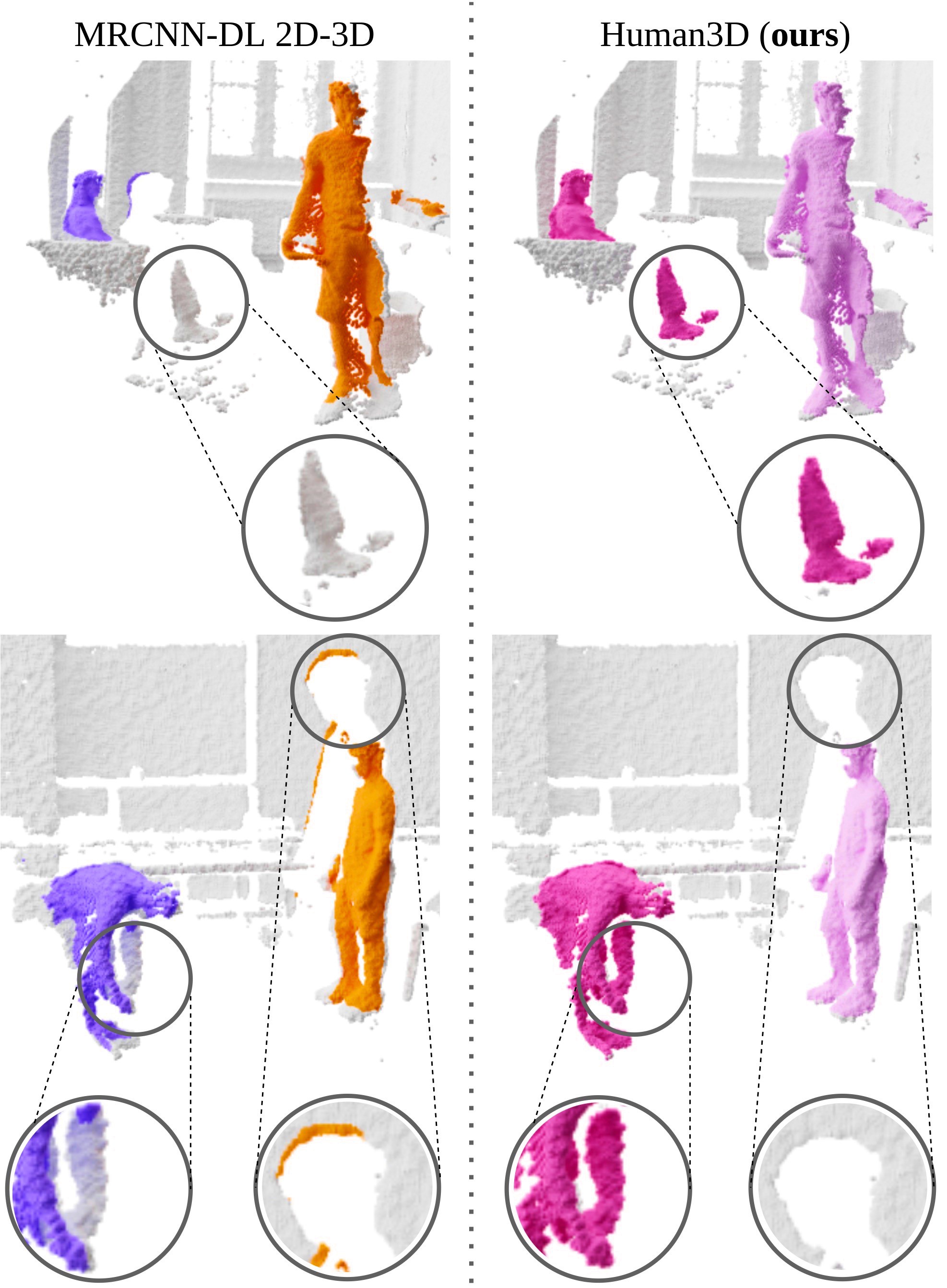}
    \caption{\textbf{Failure cases of the 2D baseline.}
    The first row shows a typical error of the Mask-RCNN baseline.
    The sofa occludes most of the human resulting in an incomplete human mask.
    In addition, the second example shows that small errors at the boundaries in 2D lead to incorrectly predicted 3D points projected far away.
    }
    \label{fig:2d_failure_cases}
\end{figure}

In this section, we provide further implementation details about the \textit{Mask-RCNN+DeepLabv3 2D-3D} baseline. This baseline closely follows the approach from \cite{egobody}. The human semantic segmentation is obtained by applying a pretrained DeepLabv3\,\cite{deeplab} to the Kinect RGB image. To obtain human instances, a pretrained Mask-RCNN is applied\,\cite{maskrcnn}.
The final 2D human instance masks are then obtained by taking the intersection of the instance and semantic masks.
These are then projected onto the 3D point cloud.
The results are shown in \reftab{2d_baseline}.
Both Mask3D\,\cite{mask3d} and our method \name{} outperform the baseline even without relying on color information, specifically on the AP$^H$ metric which is more sensitive to inaccurate mask predictions.
For both, we show the results of the models trained only on EgoBody as well as additionally pretrained on our synthetic data followed by finetuning on EgoBody, whereas the baseline is pretrained on much larger image datasets.
The error cases are due to small mistakes in 2D at the boundary of a person which project to points far away in 3D. The baseline also has more difficulties to handle occlusions. Both scenarios show the advantage of directly operating on 3D data. We illustrate these cases in Fig.~\ref{fig:2d_failure_cases}. %
\begin{table}[!ht]
\centering
\setlength{\tabcolsep}{4pt}
\resizebox{\columnwidth}{!}{
\begin{tabular}{lccc}
\toprule
    Model & Input & AP$^H$ & AP$^H_{50}$ \\
\midrule
    MRCNN-DL 2D-3D  & RGB                         & 61.3 & 97.3\\
\arrayrulecolor{black!10}\midrule\arrayrulecolor{black}

    Mask3D \emph{(no pretraining)}  & Geo. only      & 89.4 & 95.4\\
    Mask3D \emph{(pretrain.+finetune)}  & Geo. only & 95.6 & 98.7\\
    \arrayrulecolor{black!10}\midrule\arrayrulecolor{black}

    Human3D \emph{(no pretraining)}  & Geo. only      & 90.5 & 95.2\\
    Human3D \emph{(pretrain.+finetune)}  & Geo. only & \textbf{99.1} & \textbf{100}\\
\bottomrule
\end{tabular}
}
    \caption{\textbf{Comparison to image baseline.} 3D instance segmentation scores on EgoBody test set. See also Tab.~3 in  main paper.}
    \label{tab:2d_baseline}
\end{table}

\iccvsection{Qualitative Results}

\begin{figure*}[!t]
  \centering
   \includegraphics[draft=False, width=1\linewidth]{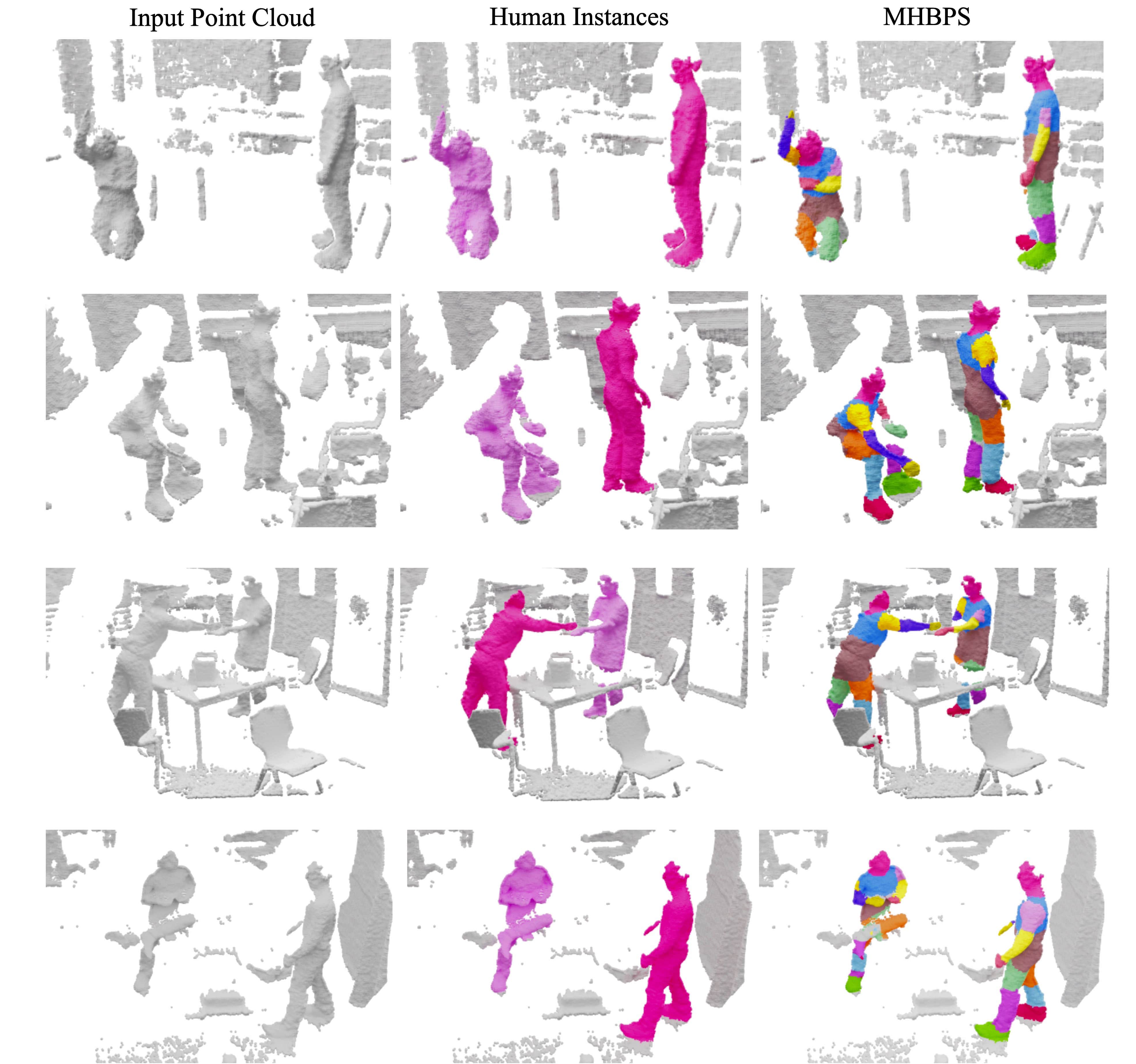}%
{
\\
\footnotesize
\ColorMapCircle{head}\,Head
\ColorMapCircle{rightArm}\,RightArm
\ColorMapCircle{leftArm}\,LeftArm
\ColorMapCircle{rightForeArm}\,RightForeArm
\ColorMapCircle{leftForeArm}\,LeftForeArm
\ColorMapCircle{rightHand}\,RightHand
\ColorMapCircle{leftHand}\,LeftHand
\ColorMapCircle{torso}\,Torso
\ColorMapCircle{hips}\,Hips

\ColorMapCircle{rightUpLeg}\,RightUpLeg
\ColorMapCircle{leftUpLeg}\,LeftUpLeg
\ColorMapCircle{rightLeg}\,RightLeg
\ColorMapCircle{leftLeg}\,LeftLeg
\ColorMapCircle{rightFoot}\,RightFoot
\ColorMapCircle{leftFoot}\,LeftFoot
}
\caption{
\textbf{Qualitative Results on EgoBody Test Set.} 
We show additional qualitative results of \name{} on the EgoBody test set.
\name{} produces strong results even for humans in challenging poses, closely interacting or occluded by scene objects.
The last row shows a failure case where \name{} predicts wrong body-parts for crossed legs.
\vspace{0cm}
}
\label{fig:qualitative_results_supp}
\end{figure*}

\parag{EgoBody Test.}
In \reffig{qualitative_results_supp}, we show additional qualitative results of \name{} on the EgoBody test set.

\begin{figure*}[!t]
  \centering
   \includegraphics[draft=False, width=1.0\linewidth]{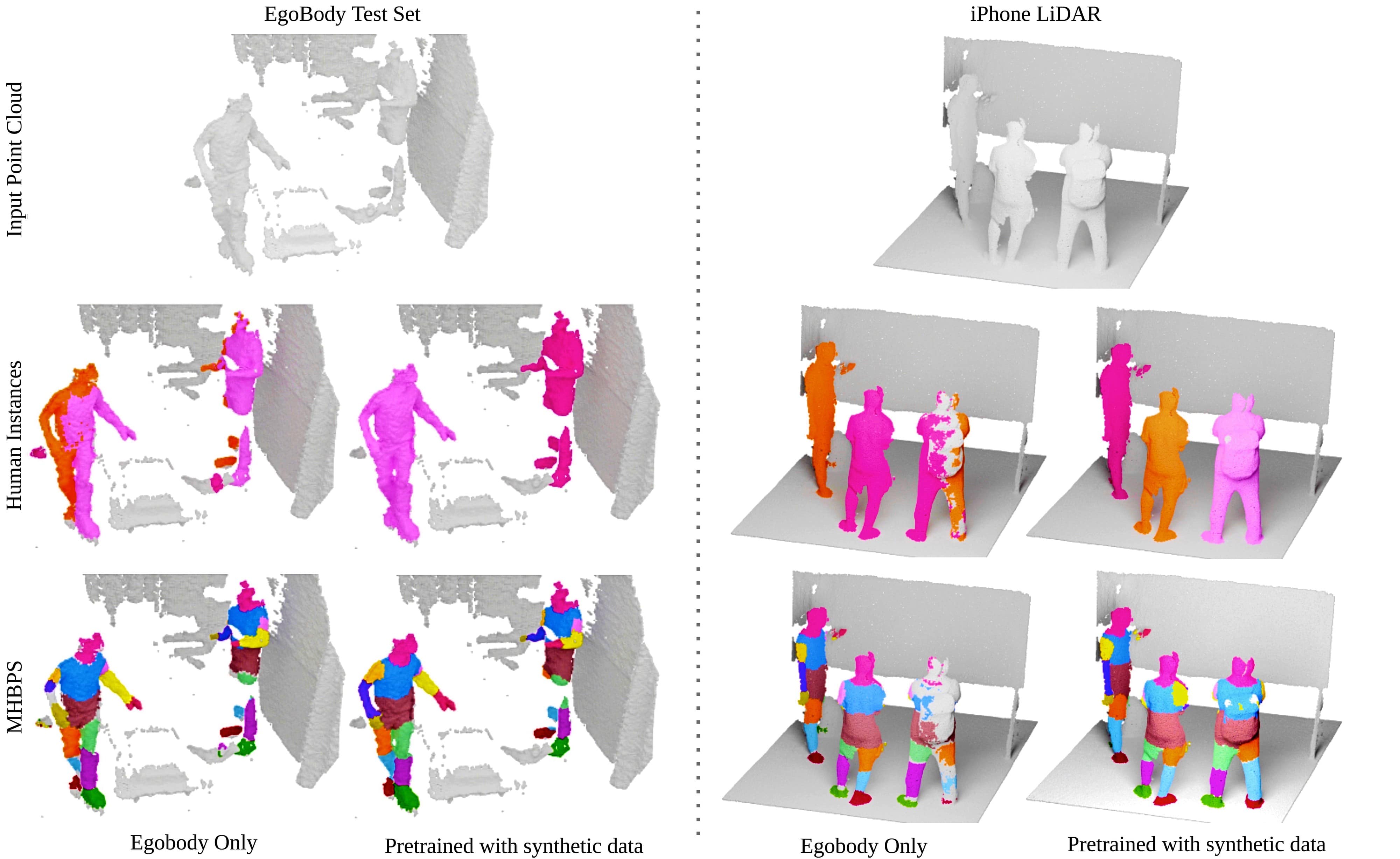}
{
\vspace{5px}\\
\footnotesize
\ColorMapCircle{head}\,Head
\ColorMapCircle{rightArm}\,RightArm
\ColorMapCircle{leftArm}\,LeftArm
\ColorMapCircle{rightForeArm}\,RightForeArm
\ColorMapCircle{leftForeArm}\,LeftForeArm
\ColorMapCircle{rightHand}\,RightHand
\ColorMapCircle{leftHand}\,LeftHand
\ColorMapCircle{torso}\,Torso
\ColorMapCircle{hips}\,Hips

\ColorMapCircle{rightUpLeg}\,RightUpLeg
\ColorMapCircle{leftUpLeg}\,LeftUpLeg
\ColorMapCircle{rightLeg}\,RightLeg
\ColorMapCircle{leftLeg}\,LeftLeg
\ColorMapCircle{rightFoot}\,RightFoot
\ColorMapCircle{leftFoot}\,LeftFoot
}
\caption{
\textbf{Pre-training with synthetic data improves upon training with EgoBody data only.}
In contrast to only training on real EgoBody data, \name{} pre-trained with synthetic data shows significantly better human instance predictions and even generalizes to scenes with more than 2 individuals.
\vspace{0cm}
}
\label{fig:pretraining_vs_only}
\end{figure*}

\begin{figure*}[!h]
  \centering
   \includegraphics[draft=False, width=1.0\linewidth]{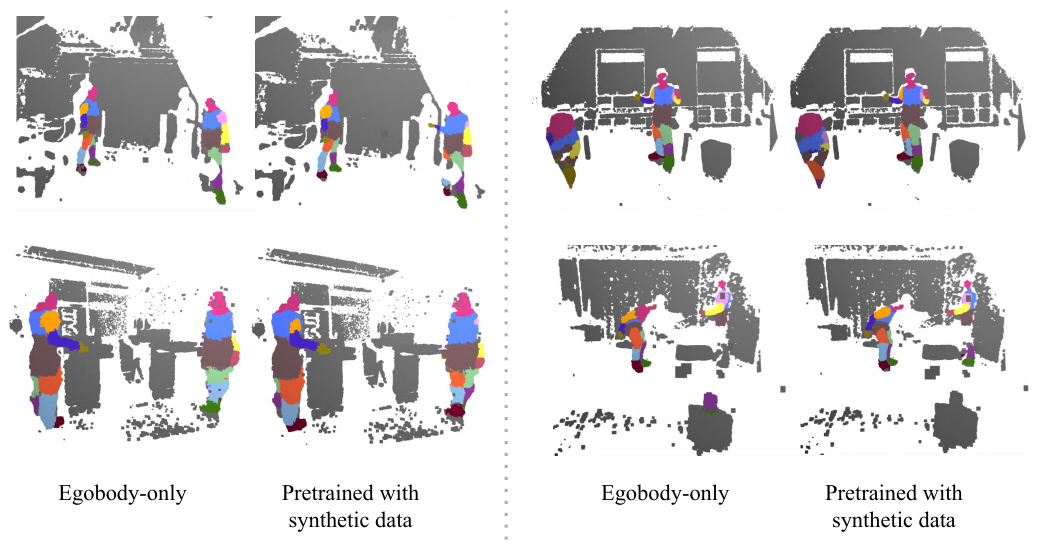}%
{
\vspace{5px}\\
\footnotesize
\ColorMapCircle{head}\,Head
\ColorMapCircle{rightArm}\,RightArm
\ColorMapCircle{leftArm}\,LeftArm
\ColorMapCircle{rightForeArm}\,RightForeArm
\ColorMapCircle{leftForeArm}\,LeftForeArm
\ColorMapCircle{rightHand}\,RightHand
\ColorMapCircle{leftHand}\,LeftHand
\ColorMapCircle{torso}\,Torso
\ColorMapCircle{hips}\,Hips

\ColorMapCircle{rightUpLeg}\,RightUpLeg
\ColorMapCircle{leftUpLeg}\,LeftUpLeg
\ColorMapCircle{rightLeg}\,RightLeg
\ColorMapCircle{leftLeg}\,LeftLeg
\ColorMapCircle{rightFoot}\,RightFoot
\ColorMapCircle{leftFoot}\,LeftFoot
}
\caption{
\textbf{Pre-training with synthetic data improves upon training with EgoBody data only.}
Model only trained with EgoBody data often confuses body parts (e.g. left leg, right leg), and struggles in the presence of occlusions. In contrast to only training on real EgoBody data, \name{} pre-trained with synthetic data shows better body-part predictions on  examples from the EgoBody test set. 
\vspace{0cm}
}
\label{fig:pretraining_vs_only_ego}
\end{figure*}

\parag{Synthetic Data Pre-Training.}
In \reffig{pretraining_vs_only} and \reffig{pretraining_vs_only_ego}, we qualitatively compare \name{} pre-trained on synthetic data with \name{} trained only on real EgoBody data.
In \reffig{pretraining_vs_only}, we observe that \name{} only trained on EgoBody data does not generalize to scenes with more than 2 individuals.
The reason for this is that the EgoBody dataset only contains scenes with less than 3 people.
When trained only on EgoBody, Human3D inevitably learns this bias and consequently fails on scenes with more than 2 people.
In contrast, our synthetic dataset consists of scenes with up to 10 people.
Human3D, pre-trained on synthetic data and fine-tuned on real EgoBody data, shows significantly better results for scenes with a larger number of people. 
In \reffig{pretraining_vs_only_ego}, we observe that pre-training with synthetic data provides robustness to occlusions and unusual poses, and results in improved multi-human body part segmentation predictions.

\clearpage

\end{document}